\documentclass[10pt,twocolumn,letterpaper]{article}

\usepackage{cvpr}              

\definecolor{cvprblue}{rgb}{0.21,0.49,0.74}
\usepackage[pagebackref,breaklinks,colorlinks,allcolors=cvprblue]{hyperref}

\usepackage{bm}
\usepackage{multirow}
\usepackage{xcolor,colortbl}         
\usepackage{algorithm}
\usepackage{algorithmic}
\usepackage{bbding}
\usepackage{xspace}

\definecolor{mygray}{RGB}{210, 210, 210}
\definecolor{mydeepgray}{RGB}{140, 140, 140}
\definecolor{myblue}{RGB}{12, 125, 118}
\definecolor{myred}{RGB}{253, 92, 155}
\definecolor{mygreen}{RGB}{79, 173, 91}

\def \myrc{\textcolor{myred}}

\def\paperID{6064} 
\def\confName{CVPR}
\def\confYear{2025}

\title{Style Quantization for Data-Efficient GAN Training}

\author{Jian Wang\thanks{Equal contribution.}, Xin Lan$^{*}$, Jizhe Zhou, Yuxin Tian, Jiancheng Lv\thanks{Corresponding author.} \\
College of Computer Science,
Sichuan University\\
Engineering Research Center of Machine Learning and Industry Intelligence, Ministry of Education
}

\begin{document}
\maketitle
\begin{abstract}
Under limited data setting, GANs often struggle to navigate and effectively exploit the input latent space. Consequently, images generated from adjacent variables in a sparse input latent space may exhibit significant discrepancies in realism, leading to suboptimal consistency regularization (CR) outcomes.
To address this, we propose \textit{SQ-GAN}, a novel approach that enhances CR by introducing a style space quantization scheme.
This method transforms the sparse, continuous input latent space into a compact, structured discrete proxy space, allowing each element to correspond to a specific real data point, thereby improving CR performance.
Instead of direct quantization, we first map the input latent variables into a less entangled ``style'' space and apply quantization using a learnable codebook. This enables each quantized code to control distinct factors of variation.
Additionally, we optimize the optimal transport distance to align the codebook codes with features extracted from the training data by a foundation model, embedding external knowledge into the codebook and establishing a semantically rich vocabulary that properly describes the training dataset.
Extensive experiments demonstrate significant improvements in both discriminator robustness and generation quality with our method.
\end{abstract}
\section{Introduction}
\label{sec:introduction}

Generative models have made significant strides in creating human-like natural language~\cite{nips/BrownMRSKDNSSAA20}, high-quality images~\cite{cvpr/KarrasLA19, cvpr/EsserRO21, iccv/PeeblesX23}, and diverse human speech~\cite{ssw/OordDZSVGKSK16} and music~\cite{corr/abs-2005-00341}. A key factor behind these achievements is the extensive amount of training data required to capture the distribution of real data. However, data collection can be challenging due to issues related to subject matter, privacy, or copyright. Consequently, training generative models with limited data has garnered increasing attention from the research community.

Recently, numerous studies~\cite{nips/ZhaoLLZ020,nips/KarrasAHLLA20,nips/JiangDWL21,cvpr/TsengJL0Y21,aaai/Cui0LZZL22,nips/Fang0S22,cvpr/Kumari0SZ22,cvpr/CuiYZLLX23} have utilized Generative adversarial networks (GANs) to train models with limited image samples, showcasing a promising future. A core issue they aim to address is the discriminator's overfitting problem when trained with limited data. The most direct solution is augmentation, including traditional~\cite{corr/abs-2006-02595,tip/TranTNNC21}, differentiable~\cite{nips/ZhaoLLZ020}, and adaptive~\cite{nips/KarrasAHLLA20} methods to augment real or generated data. However, these methods require customized manual design for different datasets, lack scalability, and limit augmentation types to a finite range~\cite{nips/KarrasAHLLA20}. Additionally, augmentations may distort the original image semantics (see Fig.~\ref{fig:semantic_sim_a}), leading to a decline in generation quality. As a complement, model regularization techniques address discriminator degradation by enhancing robustness~\cite{corr/Sangwoo2020, iclr/ZhangZOL20, aaai/Zhao0LZOZ21, cvpr/Kumari0SZ22, cvpr/CuiYZLLX23} and diversity~\cite{cvpr/Kumari0SZ22, aaai/Cui0LZZL22, cvpr/CuiYZLLX23} of the discriminator's representations, stabilizing learning dynamics during training~\cite{cvpr/TsengJL0Y21, nips/Fang0S22}. Among these, \textit{robust representations of the discriminator are paramount}.

\begin{figure}[!tbp]
\centering
\includegraphics[width=\linewidth]{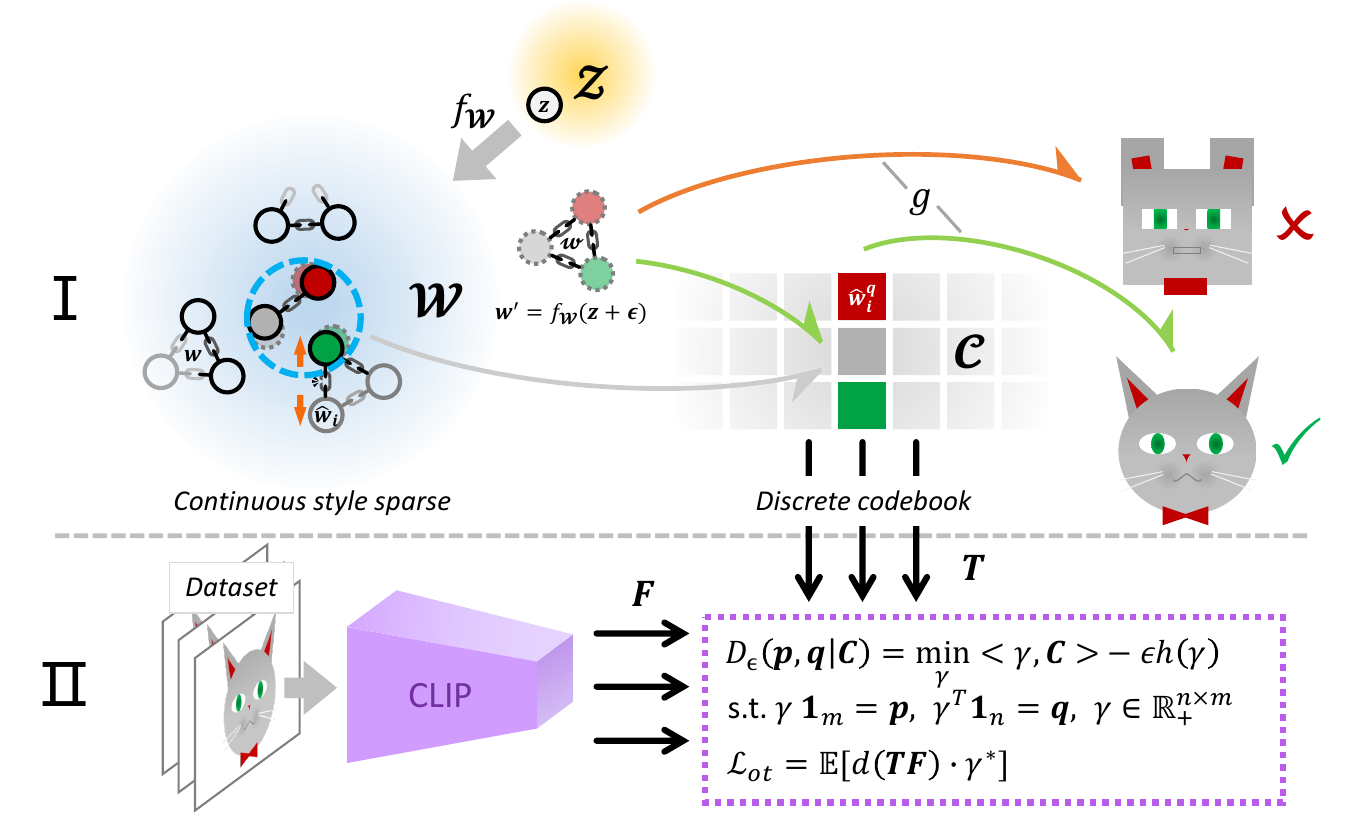}
\vspace{-0.2in}
\caption{Illustration of the proposed style quantization GAN (\textit{SQ-GAN}) framework. (\uppercase\expandafter{\romannumeral 1}) The input latent variables $\bm{z}$ are mapped to an intermediate latent space $\mathcal{W}$, which is quantized into a compact and structured proxy space $\mathcal{W}^q$ by a learnable codebook $\mathcal{C}$. The quantized codes $\bm{w}^q\in\mathcal{W}^q$ are then fed into the synthesis network to generate images. (\uppercase\expandafter{\romannumeral 2}) The codebook initialization aligns the codebook codes with features extracted from the training data, embedding external knowledge into the codebook.}
\label{fig:teaser}
\vspace{-0.2in}
\end{figure}

Consistency regularization (CR) methods~\cite{iclr/ZhangZOL20, aaai/Zhao0LZOZ21} offer an alternative perspective to encourage the discriminator to capture more robust features. These methods introduce an auxiliary term that compels the discriminator to produce consistent scores for input images (both generated and real) and their augmented versions. Particularly, \citet{aaai/Zhao0LZOZ21} propose generating samples from two adjacent latent variables, $\bm{z}$ and its perturbed version $\bm{z}+\bm{\epsilon}$ in the prior space $\mathcal{Z}$, using them as implicit augmentations to enforce consistency. This strategy circumvents the limitations of manual augmentation design and achieves impressive results. However, under limited data conditions, the effective exploitation of $\mathcal{Z}$ becomes challenging~\cite{cvpr/Kumari0SZ22, cvpr/CuiYZLLX23}. This inadequacy may result in significant disparities in authenticity between images generated from $\bm{z}$ and $\bm{z}+\bm{\epsilon}$, which, if forced to be evaluated consistently by the discriminator, could impair its effectiveness.

Our analysis highlights a key problem: how to effectively exploit the prior space $\mathcal{Z}$ with limited data such that two adjacent latent variables, $\bm{z}$ and $\bm{z}+\bm{\epsilon}$, generate diverse yet equally authentic images. The challenge resides in the inherent conflict between limited data and the comprehensive exploration of $\mathcal{Z}$. Drawing inspiration from vector quantization (VQ) techniques~\cite{nips/OordVK17, nips/RazaviOV19, cvpr/EsserRO21, nips/ZhouCLL22}, we consider compressing the representation of latent variables into a more compact and structured discrete space making each encoding more semantically rich and explicit, thus facilitating subsequent sampling to generate more diverse and consistent images.

In this paper, we introduce a style space quantization technique (\textit{SQ-GAN}) aimed at optimizing the consistency regularization of the discriminator, enhancing the robustness of its extracted features, and mitigating GAN training degradation in scenarios with limited samples.
As illustrated in Fig.~\ref{fig:teaser}, we embed the input latent variables $\bm{z}$ into an intermediate latent space $\mathcal{W}$, akin to StyleGAN~\cite{cvpr/KarrasLA19}. The intermediate variable $\bm{w}\in\mathcal{W}$ is typically interpreted as a less entangled "style," governing various high-level attributes of the generated images~\cite{cvpr/KarrasLA19}. We segment $\bm{w}$ into $s$ sub-vectors, $\{\hat{\bm{w}}_i\}_{i=1}^s$, each corresponding to different aspects or subsets of attributes. These segmented latent variables are then quantized to their nearest neighbors in a learnable codebook, enabling each quantized code $\hat{\bm{w}}_i^q$ to control distinct factors of variation.
For image generation, the quantized codes are concatenated ($\bm{w}^q = \left[\hat{\bm{w}}_1^q, \hat{\bm{w}}_2^q, \ldots\right]\in\mathcal{W}^q$) and fed into the synthesis network. 
In a limited data setting, exploring this smaller, more compact proxy space $\mathcal{W}^q$ is more feasible than directly exploring $\mathcal{W}$. This design effectively compels the intermediate latent variables $\bm{w}$ to be recombined from a limited set of distinct discrete sub-vectors $\hat{\bm{w}}_q$, allowing the generator to assign explicit meanings to each sub-vector.

Furthermore, we introduce a novel codebook initialization (CBI) method, which is crucial for capturing the intrinsic semantic structure of the training data. This method maps the codebook's codes and images from the training dataset to feature space using a foundation model, such as CLIP~\cite{icml/RadfordKHRGASAM21}. Then, it aligns these two sets of features using optimal transport distance, initializing the codebook in a manner that abstractly describes the training dataset. This alignment embeds external knowledge into the codebook, pre-establishing a semantically rich vocabulary of the training dataset.
Our contributions are fourfold:
\begin{itemize}
    \item As far as we know, this is the first work to introduce style space quantization to enhance GAN training with limited data.
    \item  A novel method for quantizing in the intermediate latent space to enhance discriminator consistency regularization.
    \item An innovative codebook initialization technique that incorporates external knowledge, pre-establishing a semantically rich vocabulary for the codebook.
    \item Comprehensive experiments demonstrating the effectiveness of our approach.
\end{itemize}

\section{Related work}
\label{sec:related}
\subsection{Advancements in GANs}
Generative adversarial networks (GANs)~\cite{nips/GoodfellowPMXWOCB14} are a groundbreaking framework designed to minimize the Jensen-Shannon (JS) divergence between real and synthetic data distributions.
Since its inception, extensive research has extended this framework by exploring a variety of adversarial loss functions that target different divergences or distances, such as the chi-squared ($\chi^2$) divergence~\cite{iccv/MaoLXLWS17}, total variation distance~\cite{iclr/ZhaoML17}, and integral probability metrics (IPMs)~\cite{icml/ArjovskyCB17,icml/SongE20}. These efforts have aimed to enhance the stability and convergence of GAN training, addressing the initial challenges associated with the original GAN framework.
The evolution of GANs has led to substantial enhancements in the quality and diversity of generated images. These advancements are attributed not only to the refinement of objective functions but also to the development of sophisticated neural network architectures~\cite{iclr/MiyatoKKY18,iclr/MiyatoK18,icml/ZhangGMO19} and robust training methodologies~\cite{nips/DentonCSF15,iccv/ZhangXL17,iclr/KarrasALL18,cvpr/Liu0BZ020}. Notable among these developments are BigGAN~\cite{iclr/BrockDS19} and StyleGAN~\cite{cvpr/KarrasLA19,cvpr/KarrasLAHLA20}, which have set new benchmarks by producing high-resolution images with intricate details and varied styles.

\subsection{GAN Training with limited data}
Training GANs with limited data poses significant challenges, primarily due to the discriminator's tendency to overfit, which can degrade the quality of generated samples~\cite{cvpr/WebsterRSJ19,iclr/GulrajaniRM19}. Data augmentation~\cite{nips/KarrasAHLLA20,corr/abs-2006-05338,iclr/ZhangZOL20,nips/ZhaoLLZ020,aaai/Zhao0LZOZ21,corr/abs-2006-02595} has emerged as a solution by enriching the dataset through transformations that diversify the original data. However, applying augmentation to GANs requires careful consideration to avoid altering the target distribution or introducing artifacts. Recent research has introduced GAN-specific augmentation methods, such as differentiable~\cite{nips/ZhaoLLZ020}, adaptive~\cite{nips/KarrasAHLLA20}, and generative~\cite{corr/abs-2006-02595} augmentations, which address these challenges effectively.

In addition to data augmentation, model regularization offers another strategy to enhance GAN training.
Various regularization techniques have been explored, including noise injection~\cite{iclr/ArjovskyB17,iclr/SonderbyCTSH17,cvpr/JenniF19}, gradient penalties~\cite{nips/GulrajaniAADC17,icml/MeschederGN18}, spectral normalization~\cite{iclr/MiyatoKKY18,iclr/MiyatoK18}, and consistency  regularization~\cite{iclr/ZhangZOL20,aaai/Zhao0LZOZ21}. Recent innovations, such as the LC regularization term~\cite{cvpr/TsengJL0Y21}, which modulates the discriminator's evaluations using two exponential moving average variables and connects to a type of \textit{f}-divergence known as LeCam divergence~\cite{ssis/Cam1986}, have proven beneficial. Additionally, DigGAN~\cite{nips/Fang0S22} addresses the gradient discrepancy between real and synthetic images by reducing this gradient gap, thus enhancing GAN performance.
Furthermore, studies have explored various methods to introduce external knowledge into the discriminator, such as using pre-trained models as additional discriminators~\cite{cvpr/Kumari0SZ22}, aligning the discriminator's features with those of a pre-trained model through knowledge distillation~\cite{cvpr/CuiYZLLX23}, or fixing the highest resolution layer of the discriminator by pre-training its parameters on a larger dataset~\cite{corr/Sangwoo2020}.

\subsection{Codebook learning for image generation}
Recent works have focused on image quantization and synthesis through codebook learning. The VQ-VAE model~\cite{nips/OordVK17} and its successor, VQ-VAE-2~\cite{nips/RazaviOV19}, enabled high-resolution image generation by capturing latent representations using discrete symbols, which facilitate efficient indexing and retrieval. \citet{icml/ChenRC0JLS20} improved quality and efficiency within these codebook constraints, while VQ-GAN~\cite{cvpr/EsserRO21} further enhanced image quality by integrating adversarial and perceptual losses. Building on these foundations, DALL-E~\cite{icml/RameshPGGVRCS21} showcased the effectiveness of discrete representations in generating images from text. The VQ-Diffusion Model~\cite{cvpr/GuCBWZCYG22} combined vector quantization with diffusion processes for superior results.

In contrast to these approaches, we introduce a novel style space quantization scheme specifically designed for GAN training with limited data. This method leverages the compact and structured properties of codebooks to effectively capture and exploit the intrinsic semantic structures present in small datasets.
\section{Preliminary}

\subsection{Generative adversarial networks}
Generative adversarial networks (GANs)~\cite{nips/GoodfellowPMXWOCB14} consist of two core components: the generator \( g \) and the discriminator \( f_D \). The generator synthesizes data samples by transforming a random noise vector drawn from a prior distribution \( p_{\bm{z}} \) (typically \( \mathcal{N}(\bm{0}, \bm{1}) \)). The discriminator, on the other hand, evaluates the real samples and these generated samples to assess their authenticity. The performance of the generator and discriminator is optimized through the following loss functions:
\begin{equation}
\label{eq:loss_adv_g}
\mathcal{L}_\text{adv}(g) = \mathbb{E}[1 - \log f_D(g(\bm{z}))],
\end{equation}
\begin{equation}
\label{eq:loss_adv_fd}
\mathcal{L}_\text{adv}(f_D) = -\mathbb{E}[\log f_D(\bm{x})] - \mathbb{E}[\log (1 - f_D(g(\bm{z})))],
\end{equation}
where $\bm{x} \in \mathbb{R}^{h \times w \times c}$ represents real data, and $h$, $w$, and $c$ denote the height, width, and channel dimensions, respectively.

\subsection{Consistency regularization}
Recent studies~\cite{iclr/ZhangZOL20} have introduced a latent consistency regularization (CR) term to ensure that the discriminator's predictions for two images generated from a latent variable $\bm{z}$ and its perturbed version $\bm{z} + \bm{\epsilon}$ remain consistent:
\begin{equation}
\label{eq:zCR_f}
\mathcal{L}_\text{cr}(f_D) = \lambda_{f_D}\mathbb{E}\left[\|f_D(g(\bm{z}))-f_D(g(\bm{z} + \bm{\epsilon}))\|^2\right].
\end{equation}
To enhance the generator's sensitivity to variations in the prior distribution, thereby increasing the diversity of generated samples, CR additionally imposes a constraint on the generator:
\begin{equation}
\label{eq:zCR_g}
\mathcal{L}_\text{cr}(g) = -\lambda_g\mathbb{E}\left[\|g(\bm{z})-g(\bm{z} + \bm{\epsilon})\|^2\right],
\end{equation}
where $\bm{\epsilon} \sim \mathcal{N}(\bm{0}, \sigma\mathbf{I})$, $\sigma$ represents the perturbation strength, set to $\sigma = 0.1$ in the experiments, and $\lambda_{f_D}$ and $\lambda_g$ are hyperparameters that balance the contributions of the respective terms.

\subsection{Challenges in CR}
Conceptually, reducing the discriminator's sensitivity to perturbations in the input latent variable to promote invariance to distortions can enable the extraction of more robust features. However, under limited data conditions, the exploration of the generator's input latent space may be insufficient. This insufficiency could result in a discontinuous mapping from latent variables to data points, leaving portions of the latent space underpopulated. Consequently, it is not guaranteed that adjacent latent variables $\bm{z}$ and $\bm{z} + \bm{\epsilon}$ will consistently map to corresponding data points, potentially degrading the effectiveness of CR.

As shown in Fig.~\ref{fig:semantic_sim_b}, the sparse nature of the input latent space may lead to significant differences in the realism of generated images $g(\bm{z})$ and $g(\bm{z} + \bm{\epsilon})$. In such cases, if the discriminator is forced to evaluate these images with consistent criteria, it may encounter difficulties, leading to suboptimal results in CR. To mitigate this issue, we propose a novel approach that incorporates intermediate latent space quantization.

\begin{figure*}[!tbp]
\centering
\includegraphics[width=0.8\linewidth]{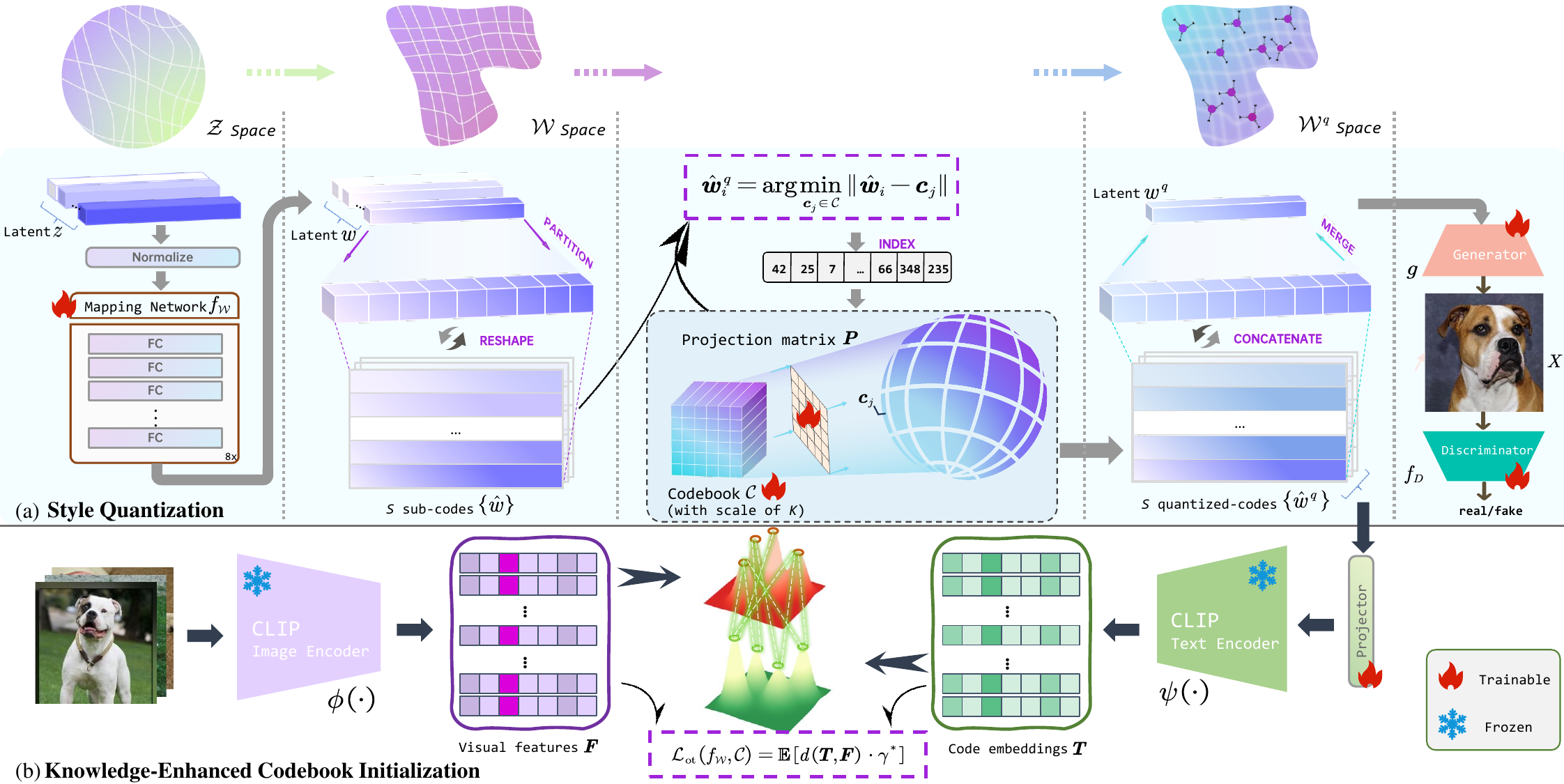}
\vspace{-0.1in}
\caption{The overall framework of SQ-GAN. (a) \textbf{Style quantization.} For a batch of intermediate variables $\bm{w}$, we apply an intermediate latent space quantization technique, segmenting and quantizing them using a hyperspherical codebook. The concatenated discrete codes are directly fed into the generator for image synthesis. (b) \textbf{Knowledge-enhanced codebook initialization.} We perform an alignment strategy grounded in optimal transport distance to embed semantic knowledge from foundation models into the codebook. The transformation and transmission of latent variables and features occur within the corresponding latent space, as illustrated at the \textbf{top} of the figure. Our framework constructs a vocabulary-rich codebook, which ensures that the entries within the codebook adequately represent a diverse and compact set of image features, suitable for image generation in limited data scenarios.}
\label{fig:framework}
\vspace{-0.2in}
\end{figure*}

\section{Style Quantization}
Our goal is to provide GANs with a more compact and structured input latent space, enabling more feasible and effective exploitation of the input latent space under limited data conditions, thereby enhancing the performance of the consistency regularization (CR). We propose an intermediate latent space quantization technique that segments the intermediate latent variables and quantizes them into a learnable codebook. To ensure each codebook entry explicitly controls different factors of variation, we introduce uniformity constraints. Finally, we present a novel quantization-based CR method.

\subsection{Style space}
In the generator architecture, the input latent variable $\bm{z} \in \mathcal{Z}$ is mapped to a less entangled intermediate latent space $\mathcal{W}$ using a non-linear mapping function \( f_\mathcal{W}: \mathbb{R}^{d_{\bm{z}}} \to \mathbb{R}^{d_{\bm{w}}} \), akin to StyleGAN~\cite{cvpr/KarrasLA19}. Here, $d_{\bm{z}}$ and $d_{\bm{w}}$ denote the dimensions of the input latent variable $\bm{z}$ and the mapped intermediate latent variable $\bm{w} \in \mathcal{W}$, respectively.
The intermediate latent variable $\bm{w}$ influences each layer of the subsequent synthesis network $g$ via affine transformations, effectively functioning as "style" to control various high-level attributes of the generated images.
Therefore, we term the intermediate latent space $\mathcal{W}$ as the \textit{style space}.

\subsection{Proxy space}
To establish an ideal mapping between the style space $\mathcal{W}$ and the data points, we propose a novel style space quantization method, termed style quantization GAN (\textit{SQ-GAN}), to construct a compact, structured, and discrete \textit{proxy space} $\mathcal{W}^q$ that better fits limited datasets.

As illustrated in Fig.~\ref{fig:framework}a, to achieve finer-grained control over the feature space and capture more detailed information, we partition each intermediate latent variable $\bm{w}$ into $s$ sub-vectors $\{\hat{\bm{w}}_i\}_{i=1}^s \subset \mathbb{R}^{s \times (d_{\bm{w}}/s)}$. We set $d_{\bm{w}}/s = 4$ to balance the trade-off between the number of partitions and image quality. Each sub-vector $\hat{\bm{w}}_i$ is optimized to manage specific details or attributes. Subsequently, we quantize each sub-vector $\hat{\bm{w}}_i$ to the nearest item in a learnable codebook $\mathcal{C}$, resulting in a quantized sub-vector $\hat{\bm{w}}_i^q$:
\begin{equation}
\hat{\bm{w}}_i^q=\left(\text{arg}\min_{\bm{c}_j\in\mathcal{C}}\lVert\hat{\bm{w}}_i-\bm{c}_j \rVert\right)\in\mathbb{R}^{(d_{\bm{w}}/s)},
\end{equation}
where $\bm{c}_j$ is the $j$-th entry in the codebook $\mathcal{C} \subset \mathbb{R}^{k \times (d_{\bm{w}}/s)}$, and $k$ denotes the number of codebook entries. This approach represents each original $\bm{w}_i$ as a set of $s$ quantized discrete representations $\{\hat{\bm{w}}_i^q\}_{i=1}^s$. Finally, the synthesis network $g$ generates a high-quality image $\tilde{\bm{x}} = g(\hat{\bm{w}}^q)$ from the latent code $\bm{w}^q \in \mathcal{W}^q$, formed by concatenating the quantized codes $\{\hat{\bm{w}}_i^q\}_{i=1}^{s}$:
\begin{equation}
\bm{w}^q = \left[\hat{\bm{w}}_1^q, \hat{\bm{w}}_2^q, \ldots, \hat{\bm{w}}_s^q\right]\in\mathbb{R}^{d_{\bm{w}}}.
\end{equation}
Thus, the proxy space $\mathcal{W}^q$ is defined as the Cartesian product of the $s$ codebooks, $\mathcal{W}^q = \mathcal{C}^1 \times \mathcal{C}^2 \times \ldots \times \mathcal{C}^s$.

\subsection{Loss function}
We define the following loss functions to train our model end-to-end:
\begin{equation}
\mathcal{L}(g, f_\mathcal{W}, \mathcal{C}, \bm{P}) = \mathcal{L}_\text{adv}(g) + \lambda_\text{sq}(\mathcal{L}_\text{sq}(f_\mathcal{W}, \mathcal{C}) + \mathcal{L}_\text{uf}(\mathcal{C}, \bm{P})),
\end{equation}
\begin{equation}
\mathcal{L}(f_D) = \mathcal{L}_\text{adv}(f_D) + \lambda_\text{qcr}\mathcal{L}_\text{qcr}(f_D),
\end{equation}
where $\lambda_\text{sq}$ and $\lambda_\text{qcr}$ are hyperparameters that balance the contributions of the corresponding terms. In our experiments, we set $\lambda_\text{sq} = 0.01$ and $\lambda_\text{qcr} = 0.01$.

\paragraph{Adversarial losses $\mathcal{L}_\text{adv}(g)$ and $\mathcal{L}_\text{adv}(f_D)$} The adversarial losses, defined in Eq.~(\ref{eq:loss_adv_g}) and Eq.~(\ref{eq:loss_adv_fd}), are designed to optimize the generator $g$ and the discriminator $f_D$.

\paragraph{Style quantization loss $\mathcal{L}_\text{sq}(g)$}
The style quantization loss~\cite{nips/OordVK17} optimizes the distance between the split latent variable $\hat{\bm{w}}$ and the corresponding codebook representation $\mathcal{C}$:
\begin{equation}
\label{eq:loss_sq}
\mathcal{L}_\text{sq}(f_\mathcal{W}, \mathcal{C}) = \|\mathrm{sg}\left(\hat{\bm{w}}\right)-\bm{c}\|_2^2+\beta\|\hat{\bm{w}}-\mathrm{sg}\left(\bm{c}\right)\|_2^2,
\end{equation}
where $\mathrm{sg}(\cdot)$ is the stop-gradient operator, and the second term, scaled by $\beta$, is a so-called ``commitment loss'' that prevents the learned codes in $\mathcal{C}$ from growing arbitrarily.
To handle the non-differentiable operation in Eq.~(\ref{eq:loss_sq}), we employ the straight-through gradient estimator, which allows gradients from the synthesis network $g$ to flow directly to the mapping network $f_\mathcal{W}$ during backpropagation.

\paragraph{Uniformity regularization $\mathcal{L}_\text{uf}(\mathcal{C},\bm{P})$}
To prevent codebook collapse during limited data training, we impose a uniformity constraint on the codebook $\mathcal{C}$. We first project the codebook $\mathcal{C}$ onto a unit hypersphere using a learnable projection matrix $\bm{P}$ and $l_2$ normalization, such that $\bar{\bm{c}}_i = \bm{P}\bm{c}_i / \|\bm{P}\bm{c}_i\|_2$. The uniformity problem on the unit hypersphere is characterized by minimizing the total pairwise potential based on a specific kernel function~\cite{Cook_1973, borodachov2019discrete}. Following~\cite{icml/0001I20}, we employ the Radial Basis Function (RBF) kernel, which can encapsulate a broad spectrum of kernel functions~\cite{Cohn_Kumar_2006}:
\begin{equation}
\label{eq:uniform}
\mathcal{L}_\text{uf}(\mathcal{C}, \bm{P}) = \log\mathbb{E}\left[ \exp\left({-t\|\bar{\bm{c}}_i-\bar{\bm{c}}_j\|_{2}^{2}}\right) \right],
\end{equation}
where $t > 0$ is a fixed scale parameter. This regularization promotes uniform distribution of feature codes on the unit hypersphere, ensuring diverse and mean-explicit codes within the codebook throughout training.

\paragraph{Quantization-based consistency regularization $\mathcal{L}_\text{qcr}$}
In quantization-based CR, we enforce consistency in the discriminator's predictions for images generated from quantized intermediate latent variables $\bm{w}^q$ and their perturbed versions. The perturbed intermediate latent variable is obtained by adding Gaussian noise $\bm{\epsilon}$ (same as CR) to the original input latent variable $\bm{z}$, then mapping it to the intermediate latent variable $\bm{w}' = f_\mathcal{W}(\bm{z}+\bm{\epsilon})$, and quantizing it via the codebook $\mathcal{C}$ to get $\bm{w}'^q$. The quantization-based CR loss is defined as:
\begin{equation}
\label{eq:qcr}
\mathcal{L}_\text{qcr}(f_D) = \mathbb{E}\left[\|f_D(g(\bm{w}^q)) - f_D(g(\bm{w}'^q))\|^2\right].
\end{equation}
This form of regularization ensures that even minor perturbations in the latent space, leading to different quantized representations, result in consistent evaluations by the discriminator.

\section{Knowledge-Enhanced Codebook Initialization}
\label{sec:codebook_init}

Our approach aims to ensure that the codes within the codebook properly represent various image features, subsequently using their combinations to correspond to specific data points. However, deriving such codes from a limited training dataset is a non-trivial task. To address this, we propose an alignment strategy based on optimal transport distance for codebook initialization (CBI). This method leverages the semantic knowledge embedded in foundation models to construct a richer vocabulary for the training dataset.

\subsection{Feature extraction}
As depicted in Fig.~\ref{fig:framework}b, we begin by utilizing a foundation model to encode both the images in the dataset and the codes (quantized sub-vectors) in the codebook into their respective feature spaces. Given that the initialization of the codebook is essentially a pairing task, we leverage CLIP~\cite{icml/RadfordKHRGASAM21}, a widely adopted large-scale vision-language pre-trained model. The extensive pretraining of CLIP enables the exploration of semantic correspondences between visual and linguistic modalities.

\paragraph{Image encoding}
In the image encoding process, we directly employ CLIP's visual encoder to extract image features. Specifically, given a batch of $n$ real images $\{\bm{x}_i\}_{i=1}^n$ from a limited dataset, we extract the features for each image, $\bm{f}_i=\phi(\bm{x}_i)\in\mathbb{R}^{l\times d_e}$, using CLIP's visual encoder $\phi$. Here, $l$ represents the image token sequence length, and $d_e$ denotes the embedding dimension of the image feature. Consequently, we obtain a set of image features $\bm{F}=\{\bm{f}_i\}_{i=1}^n\subset\mathbb{R}^{n\times l\times d_e}$.

\paragraph{Code embedding}
Consider $m$ latent variables $\{\bm{z}_i\}_{i=1}^m$ sampled from the prior distribution. These are first mapped to the style space $\mathcal{W}$ and then quantized into the discrete proxy space $\mathcal{W}^q$. To align with the Transformer architecture, the quantized sub-vectors $\{\hat{\bm{w}}^q_i\}_{i=1}^m\subset\mathbb{R}^{m\times s\times(d_{\bm{w}}/s)}$ are passed through a trainable MLP layer, transforming them into variables with the same embedding dimension as the input tokens. These transformed variables are then processed by a specific Transformer structure. In our implementation, we utilize CLIP’s text encoder, $\psi$, which offers a critical advantage: the features derived from both the visual and text encoders are already closely aligned, thereby reducing feature discrepancy and accelerating training convergence. Thus, we map the $m$ quantized discrete codes $\{\hat{\bm{w}}^q_i\}_{i=1}^m$ to the feature space, resulting in the feature set $\bm{T}=\{\bm{t}_i\}_{i=1}^m\subset\mathbb{R}^{m\times s\times d_e}$.

It is important to highlight that we treat these discrete codes as analogous to text encodings composed of words from a vocabulary. Consequently, the initialized codebook can be viewed as a vocabulary that abstractly describes the image features within the dataset.

\subsection{Aligning via optimal transport distance}
Given that the images and latent variables in each batch are unpaired, and considering the differences in semantic structure and interpretation between the image feature $\bm{F}$ and the code embedding $\bm{T}$, we employ the optimal transport (OT) distance as an effective metric to quantify their discrepancy.

\paragraph{Optimal transport distance}
Consider two distributions of features generated by a neural network: $\{ \textbf{f}^x_i \}_{i=1}^n$, with corresponding weights (marginal distribution) $\mathbf{p} = \{ p_i \}_{i=1}^n$, and $\{ \textbf{f}^y_j \}_{j=1}^m$, with weights $\mathbf{q} = \{ q_j \}_{j=1}^m$. The objective is to determine a transport plan $\gamma \in \mathbb{R}^{n \times m}$ that minimizes the total transportation cost of mapping one feature distribution to the other. The OT problem is formally defined as:
\begin{equation}
\label{eq:naive_ot}
\begin{aligned}
D(\bm{p}, \bm{q} | \textbf{C}) &= \min_{\gamma \in \mathbb{R}^{n \times m}} \left<\gamma, \textbf{C}\right> \\
&= \min_{\gamma \in \mathbb{R}^{n \times m}} \sum_{i=1}^n \sum_{j=1}^m \gamma_{ij} c(\textbf{f}^x_i, \textbf{f}^y_j), \\
\text{subject to} \quad &\sum_{j=1}^m \gamma_{ij} = p_i \quad \forall i \in \{1, \ldots, n\}, \\
&\sum_{i=1}^n \gamma_{ij} = q_j \quad \forall j \in \{1, \ldots, m\},
\end{aligned}
\end{equation}
where $c(\textbf{f}^x_i, \textbf{f}^y_j)$ represents the cost of transporting a unit of mass from feature $\textbf{f}^x_i$ to $\textbf{f}^y_j$, and $\left<\gamma, \textbf{C}\right>$ denotes the Frobenius inner product of $\gamma$ and the cost matrix $\textbf{C}$.
Directly solving this problem is computationally intensive, especially in high-dimensional feature spaces. To mitigate this, we employ the Sinkhorn divergence algorithm ~\cite{nips/Cuturi13}, which incorporates an entropic regularizer to accelerate computation. Detailed derivations can be found in Appendix \myrc{A}.

\paragraph{Alignment}
Given the marginal distributions $\mathbf{p}$ and $\mathbf{q}$ of discrete code embeddings $\bm{T}=\{\bm{t}_i\}_{i=1}^m\subset\mathbb{R}^{m\times s\times d_e}$ and visual features $\bm{F}=\{\bm{f}_i\}_{i=1}^n\subset\mathbb{R}^{n\times l\times d_e}$, alongside the corresponding cost matrix $\mathbf{C}$, the optimal transport plan $\gamma^*$ is obtained by solving equation \eqref{eq:naive_ot}. For simplicity, we initialize $\mathbf{p}$ and $\mathbf{q}$ as uniform distributions, i.e.,
\begin{equation}
    p_i = \frac{1}{s}, \quad \forall i \in \{1, \ldots, s\}, \quad q_j = \frac{1}{l}, \quad \forall j \in \{1, \ldots, l\}.
\end{equation}
The cost matrix $\mathbf{C}$ is computed by:
\begin{equation}
C_{j,k} = c(j,k) := d(t_{\cdot j}, f_{\cdot k}) \quad 1 \leq j \leq s, \quad 1 \leq k \leq l,
\end{equation}
where $d(\cdot, \cdot)$ denotes the distance metric (e.g., Euclidean distance or cosine distance) serving as the transport cost function $c$ between corresponding components of feature $\bm{t}_i$ and $\bm{f}_i$.

After computing $\gamma^*$, we define the optimal transport loss between the two features, $\bm{T}$ and $\bm{F}$, to achieve the alignment:
\begin{equation}
\mathcal{L}_\text{ot}(f_\mathcal{W}, \mathcal{C}) = {\mathbb{E}}\left[ d(\bm{T}, \bm{F}) \cdot \gamma^* \right].
\end{equation}
During the codebook initialization phase, we optimize the weighted sum of the OT loss $\mathcal{L}_\text{ot}$ along with the two additional loss terms \eqref{eq:loss_sq} and \eqref{eq:uniform}. The overall loss function is defined as:
\begin{equation}
\mathcal{L}(f_\mathcal{W}, \mathcal{C}, \bm{P}) = \mathcal{L}_\text{sq}(f_\mathcal{W}, \mathcal{C}) + \mathcal{L}_\text{uf}(\mathcal{C}, \bm{P}) + \mathcal{L}_\text{ot}(f_\mathcal{W}, \mathcal{C}).
\end{equation}

\section{Experiments}
\label{sec:experiments}

\paragraph{Experimental settings}
To evaluate the performance of our proposed SQ-GAN, we conduct experiments on four widely used datasets in the image generation under limited-data setting research field: Oxford-Dog (from Oxford-IIIT pet dataset ~\cite{cvpr/ParkhiVZJ12}, detailed in Appendix \myrc{B}), Flickr-Faces-HQ (FFHQ) ~\cite{cvpr/KarrasLA19}, MetFaces ~\cite{nips/KarrasAHLLA20}, and BreCa-HAD ~\cite{springer/aksac2019brecahad}.
The datasets are resized to $256 \times 256$ pixels for training. Detailed descriptions of these datasets are provided in Appendix \myrc{B}.

We adopt the evaluation metrics: Inception Score (IS) ~\cite{nips/SalimansGZCRCC16}, Fréchet Inception Distance (FID) ~\cite{nips/HeuselRUNH17}, and Kernel Inception Distance (KID) ~\cite{iclr/BinkowskiSAG18} to evaluate our models. The official implementations of these metrics, as provided by ~\cite{cvpr/KarrasLAHLA20}, are used for all evaluations. For the formal definition of the metrics, see Appendix \myrc{B}.

\begin{table}[tbp]
\centering
\resizebox*{\linewidth}{!}{
\begin{tabular}{|lrrrr|}
\hline
\multicolumn{1}{|c}{\multirow{2}{*}{\textbf{Method}}} & \multicolumn{2}{c}{\textbf{Oxford-Dog}} & \multicolumn{2}{c|}{\textbf{FFHQ-2.5k}} \\
\cline{2-5} 
\multicolumn{1}{|c}{} & FID $\Downarrow$ & IS $\Uparrow$ & FID $\Downarrow$ & IS $\Uparrow$ \\
\hline
StyleGAN2~\cite{cvpr/KarrasLAHLA20} & 64.26 & 9.69 & 48.11 & 3.50 \\
\hline
\multicolumn{5}{|c|}{GAN loss functions} \\
\hline
+ Wasserstein~\cite{nips/GulrajaniAADC17} & 82.18 & 9.94 & 38.64 & 4.11 \\
+ LS~\cite{iccv/MaoLXLWS17} & 216.42 & 2.69 & 213.93 & 2.17 \\
+ RaHinge~\cite{iclr/Jolicoeur-Martineau19} & 38.68 & 11.14 & 32.34 & 3.96 \\
\hline
\multicolumn{5}{|c|}{Regularization act on discriminator} \\
\hline
+ LeCam~\cite{cvpr/TsengJL0Y21} & 102.87 & 8.02 & 68.85 & 3.53 \\
+ DigGAN~\cite{nips/Fang0S22} & 61.84 & 10.07 & 48.45 & 3.68 \\
+ KD-DLGAN~\cite{cvpr/CuiYZLLX23} & 54.06 & 9.73 & 49.31 & 3.67 \\
+ CR~\cite{iclr/ZhangZOL20} & 48.73 & 10.47 & 41.43 & 4.06 \\
\rowcolor{mygray}+ SQ-GAN (ours)& 36.30 & 11.52 & 25.38 & 4.17 \\
\rowcolor{mygray}+ SQ-GAN + CBI (ours)& \textbf{35.01} & \textbf{12.44} & \textbf{22.04} & \textbf{4.20} \\
\hline
\end{tabular}
}
\vspace{-0.1in}
\caption{Comprehensive comparison with SOTA methods across various settings on the Oxford-Dog and FFHQ-2.5K datasets, implemented within the StyleGAN framework, as they are all architecture-agnostic. CBI denotes our knowledge-enhanced codebook initialization process. $\Uparrow$ denotes that higher values are preferable,
while $\Downarrow$ indicates that lower values are better.}
\label{tab:comparison_comprehensive}
\vspace{-0.1in}
\end{table}

\begin{figure}[tbp]
\centering
\begin{subfigure}{0.48\linewidth}
\includegraphics[width=\linewidth]{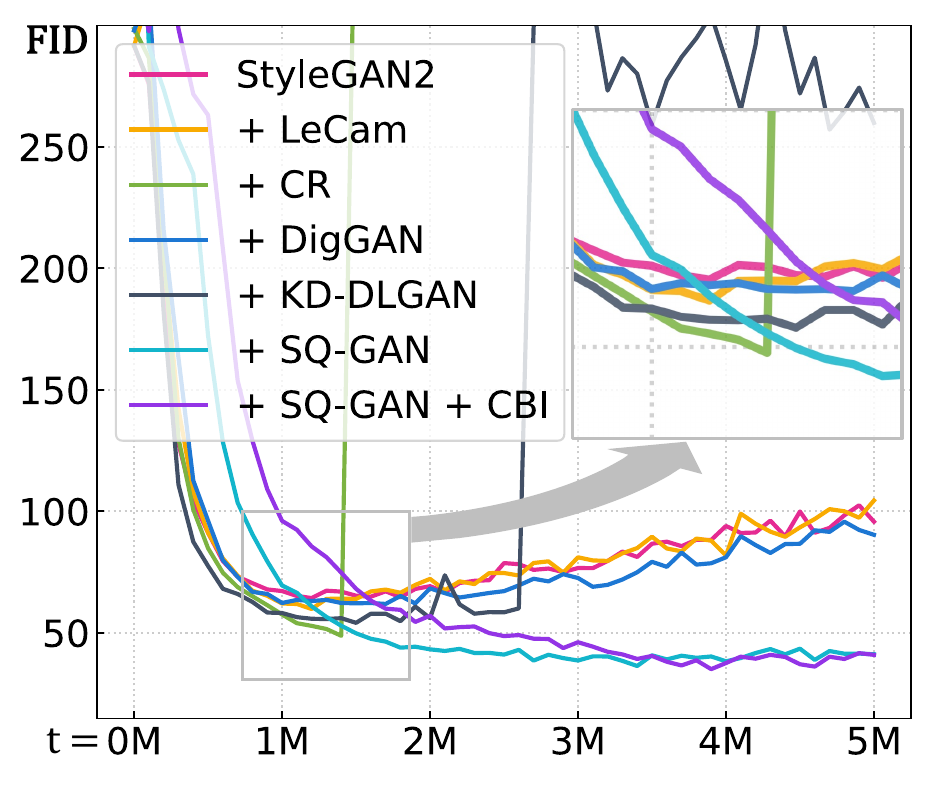}
\caption{Oxford-Dog}
\label{fig:fid_distribution_dog}
\end{subfigure}
\begin{subfigure}{0.48\linewidth}
\includegraphics[width=\linewidth]{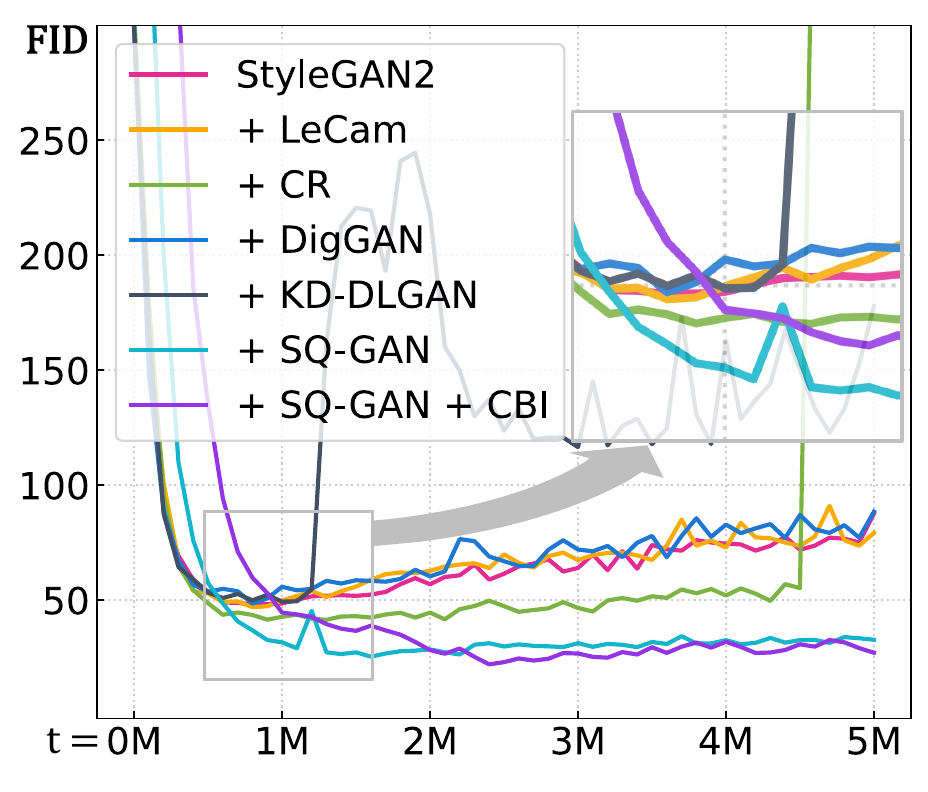}
\caption{FFHQ-2.5K}
\label{fig:fid_distribution_ffhq25}
\end{subfigure}
\vspace{-0.1in}
\caption{Evolutions of FID ($\Downarrow$) scores during training on the Oxford-Dog and FFHQ-2.5K datasets.}
\label{fig:fid_distribution}
\vspace{-0.2in}
\end{figure}

\subsection{Comparison with SOTA}
\label{sec:comparison_sota}
We conducted an extensive evaluation of our proposed SQ-GAN model against several state-of-the-art models under diverse experimental conditions. To ensure the validity of our comparisons, we used StyleGAN2~\cite{cvpr/KarrasLAHLA20} as the foundational architecture, retraining all models from scratch across all datasets. All metrics were computed using the same tool, ensuring consistency in measurement. The results of this comprehensive comparison are presented in Table \ref{tab:comparison_comprehensive} and Fig.~\ref{fig:fid_distribution}. Our baseline includes a variety of GAN loss functions, architectures, and regularization techniques. The findings demonstrate that our approach consistently outperforms the benchmarks across all metrics and datasets, showcasing its superior capability in limited data environments. Further generated images can be found in Appendix \myrc{C}.

In addition, we conducted further experiments on two particularly small datasets, MetFaces and BreCaHAD, consisting of only 1,203 and 1,750 training images, respectively. The outcomes, as shown in Table \ref{tab:performance_small_dataset}, reveal that our model effectively addresses the challenges of extremely limited data scenarios, surpassing the performance of other state-of-the-art methods. To ensure a comprehensive evaluation, we also integrated the ADA strategy~\cite{nips/KarrasAHLLA20} into our experiments. As illustrated in Table~\ref{tab:performance_small_dataset}, the combination of our method with ADA outperforms the configuration of CR with ADA, as well as all other tested configurations. This further emphasizes the effectiveness of our approach when paired with advanced augmentation techniques.

In Table \ref{tab:comparison_gan_achitecture}, we compare our approach with various GAN architectures and techniques, including DCGAN~\cite{corr/RadfordMC15} and SNGAN~\cite{iclr/MiyatoKKY18}, as well as different improvement strategies such as auxiliary rotation prediction~\cite{cvpr/ChenZRLH19}, dropout~\cite{jmlr/SrivastavaHKSS14}, and noise injection~\cite{iclr/SonderbyCTSH17}. The results demonstrate that our method significantly outperforms the traditional convolution-based DCGAN and the ResNet-based SNGAN, highlighting the effectiveness of our approach.

\begin{table}[tbp]
\centering
\resizebox*{\linewidth}{!}{
\begin{tabular}{|lrrrrrr|}
\hline
\multicolumn{1}{|c}{\multirow{2}{*}{\textbf{Method}}} & \multicolumn{3}{c}{\textbf{MetFaces}} & \multicolumn{3}{c|}{\textbf{BreCaHAD}} \\
\cline{2-7} 
\multicolumn{1}{|c}{} & FID $\Downarrow$ & IS $\Uparrow$ & KID $\Downarrow$ & FID $\Downarrow$ & IS $\Uparrow$ & KID $\Downarrow$ \\
\hline
StyleGAN2~\cite{cvpr/KarrasLAHLA20} & 53.21 & 3.16 & 0.035 & 97.06 & 3.10 & 0.095 \\
+ LeCam~\cite{cvpr/TsengJL0Y21} & 56.67 & 2.43 & 0.102 & 83.74 & 2.51 & 0.046 \\
+ CR~\cite{iclr/ZhangZOL20} & 48.89 & 3.24 & 0.029  & 80.72 & 2.92 & 0.058 \\
+ DigGAN~\cite{nips/Fang0S22} & 53.97 & 3.09 & 0.031 & 105.45 & 3.05 & 0.095 \\
+ KD-DLGAN~\cite{cvpr/CuiYZLLX23} & 54.05 & 3.11 & 0.029 & 86.25 & \textbf{3.26} & 0.083 \\
\hline
+ ADA~\cite{nips/KarrasAHLLA20} & 28.10 & 4.29 & 0.006 & 23.90 & 3.04 & 0.019 \\
+ CR + ADA & 29.91 & 4.20 & 0.008 & 22.69 & 2.81 & 0.011 \\
\hline
\rowcolor{mygray}+ SQ-GAN (ours) & 41.12 & 3.31 & 0.024 & 48.44 & 2.98 & 0.041 \\
\rowcolor{mygray}+ SQ-GAN + CBI (ours) & 35.44 & 4.02 & 0.022 & 42.42 & 2.93 & 0.037 \\
\rowcolor{mygray}+ SQ-GAN + ADA (ours) & \textbf{24.77} & \textbf{4.93} & \textbf{0.008} & \textbf{22.61} & 3.15 & \textbf{0.010} \\
\hline
\end{tabular}
}
\vspace{-0.1in}
\caption{Quantitative comparison on small (extremely limited) datasets, MetFaces and BreCaHAD.}
\label{tab:performance_small_dataset}
\end{table}

\begin{table}[t]
\centering
\resizebox*{\linewidth}{!}{
\begin{tabular}{|llrr|}
\hline
\multicolumn{2}{|l}{\textbf{Method ($64\times64$)}} & \multicolumn{1}{c}{\textbf{FID} $\Downarrow$} & \multicolumn{1}{c|}{\textbf{IS} $\Uparrow$} \\
\hline
\multirow{4}{*}{DCGAN}
& Vanilla~\cite{corr/RadfordMC15} & 107.33 & 3.23 \\
& + Auxiliary rotations~\cite{cvpr/ChenZRLH19} & 103.43 & 2.85 \\
& + Dropout~\cite{jmlr/SrivastavaHKSS14} & 77.35& 4.11 \\
& + Noise~\cite{iclr/SonderbyCTSH17} & 86.13 & 5.40 \\
\hline
\multirow{4}{*}{SNGAN}
& Vanilla~\cite{iclr/MiyatoKKY18} & 57.08 & 3.42 \\
& + Auxiliary rotations~\cite{cvpr/ChenZRLH19} & 54.85 & 3.63 \\
& + Dropout~\cite{jmlr/SrivastavaHKSS14} & 48.54 & 4.02 \\
& + Noise~\cite{iclr/SonderbyCTSH17} & 55.99 & 3.33 \\
\hline
\multirow{2}{*}{StyleGAN2}
& Vanilla~\cite{cvpr/KarrasLAHLA20} & 24.16 & 10.37 \\
& \cellcolor{mygray}+ SQ-GAN (ours)  & \cellcolor{mygray}\textbf{18.96} & \cellcolor{mygray}\textbf{11.41} \\
\hline
\end{tabular}
}
\vspace{-0.1in}
\caption{Comparison of various GAN architectures and techniques on the Oxford-Dog dataset ($64\times64$).}
\label{tab:comparison_gan_achitecture}
\vspace{-0.2in}
\end{table}

\begin{figure*}[tbp]
\centering
\begin{subfigure}{0.251\linewidth}
\includegraphics[width=\linewidth]{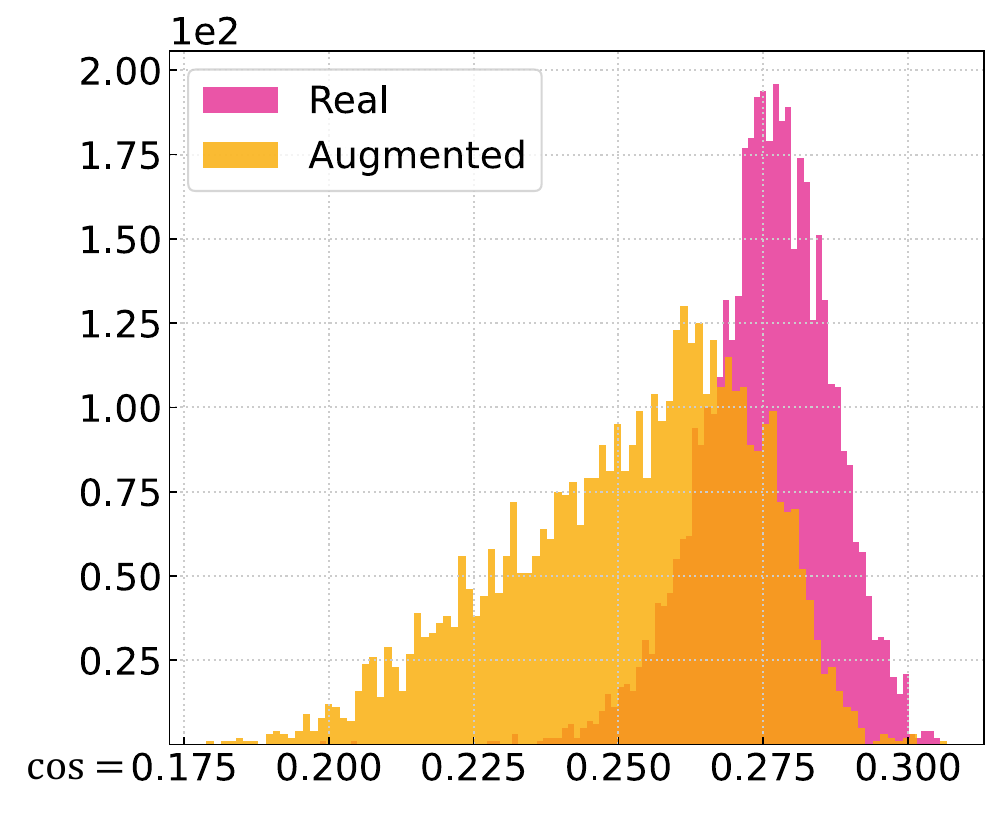}
\caption{$\Phi(\bm{x})$ and $\Phi(\text{ADA}(\bm{x}))$}
\label{fig:semantic_sim_a}
\end{subfigure}
\begin{subfigure}{0.244\linewidth}
\includegraphics[width=\linewidth]{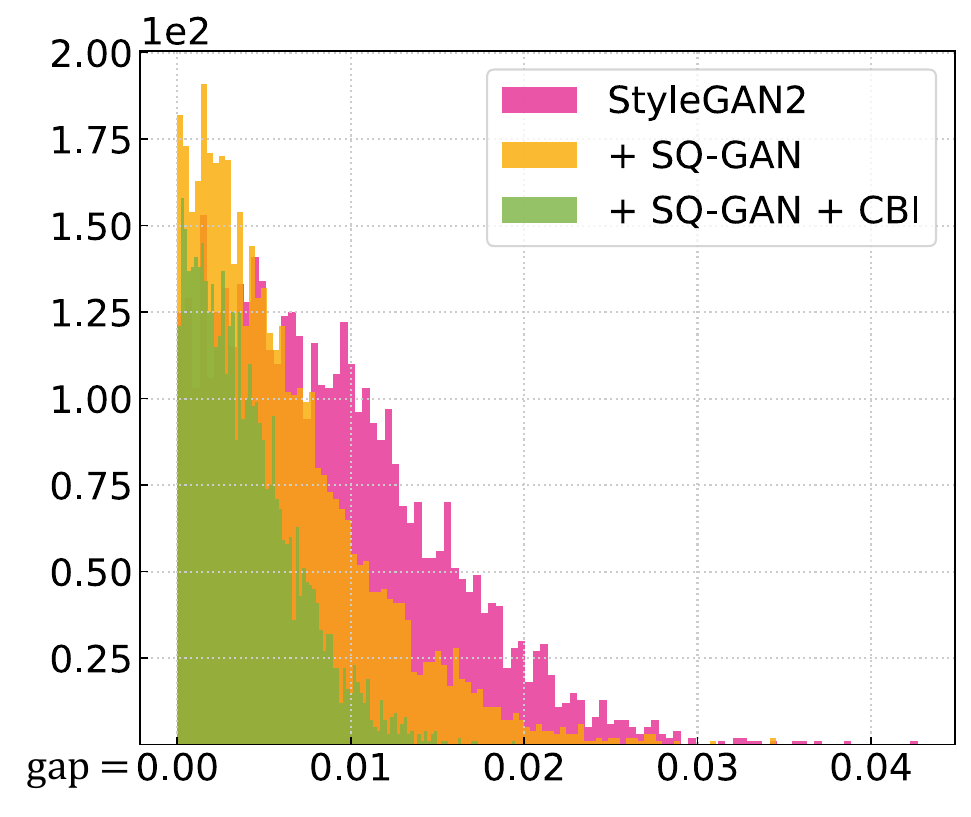}
\caption{$|\Phi(g(\bm{z})) - \Phi(g(\bm{z}+\bm{\epsilon}))|$}
\label{fig:semantic_sim_b}
\end{subfigure}
\begin{subfigure}{0.254\linewidth}
\includegraphics[width=\linewidth]{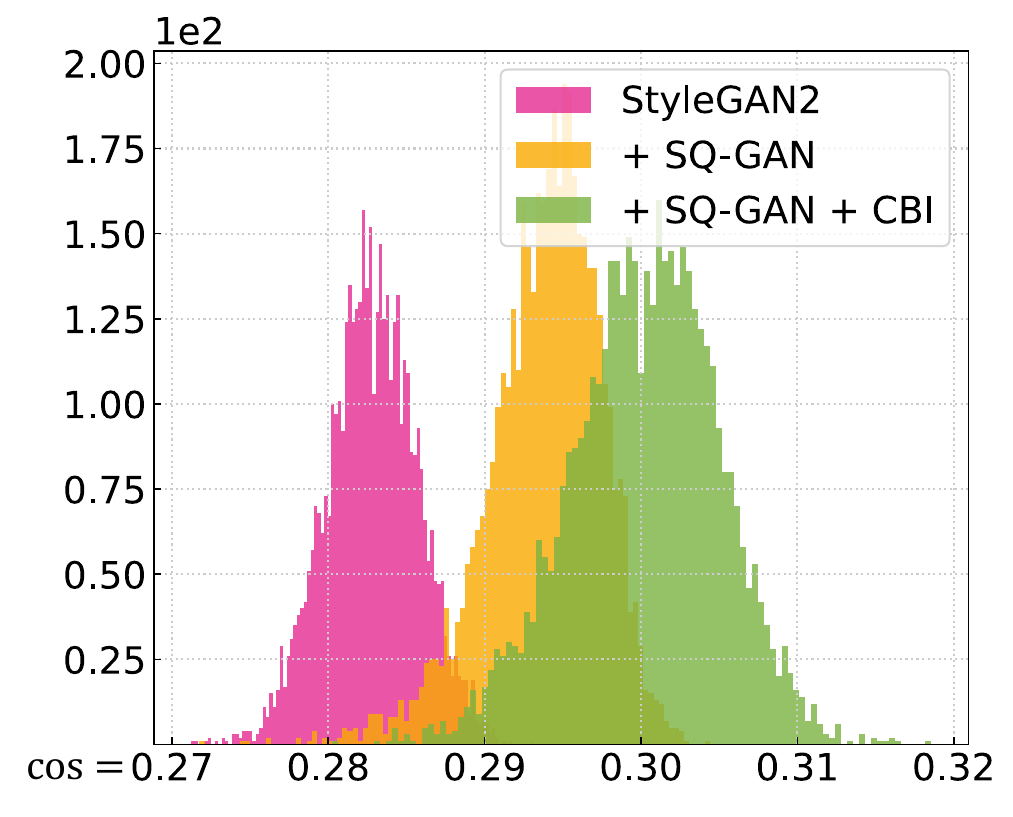}
\caption{$\Phi(g(\bm{z}))$}
\label{fig:semantic_sim_c}
\end{subfigure}
\begin{subfigure}{0.235\linewidth}
\includegraphics[width=\linewidth]{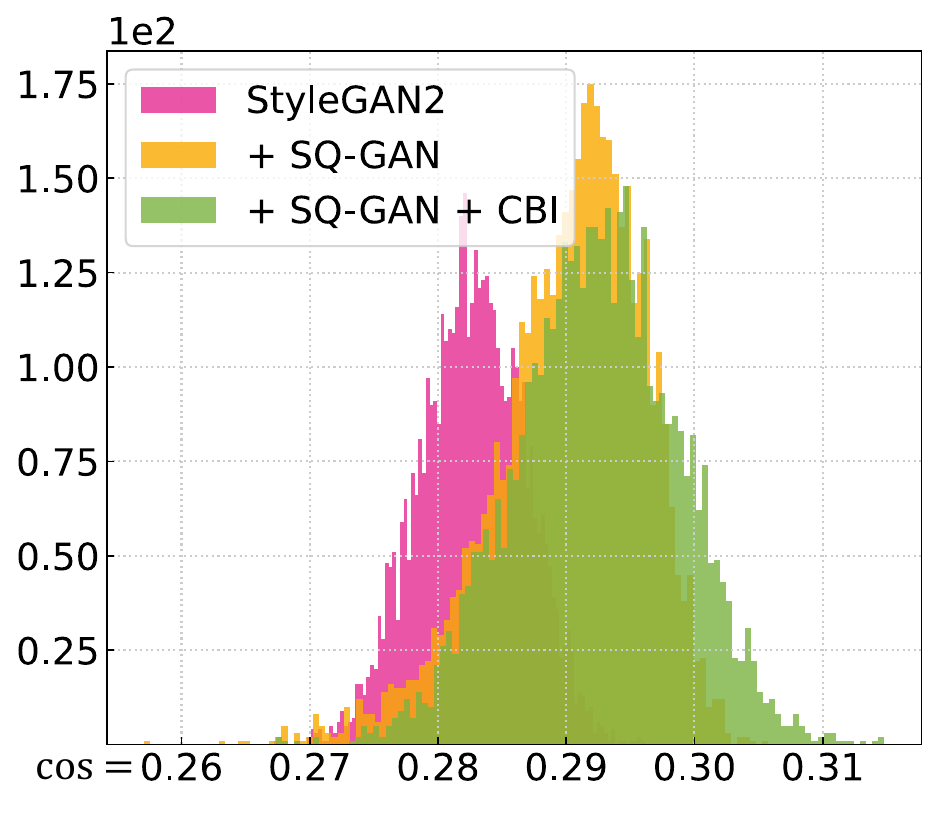}
\caption{$\Phi(g(\bm{z}+\bm{\epsilon}))$}
\label{fig:semantic_sim_d}
\end{subfigure}
\vspace{-0.25in}
\caption{Distribution of semantic similarity. We compute the cosine similarity between the features extracted from the image and semantic information from category text using the CLIP model, denoted as $\Phi(\cdot)$.}
\label{fig:semantic_sim}
\vspace{-0.2in}
\end{figure*}

\subsection{Semantic similarity analysis}
To further investigate the effectiveness of our method, we conducted a semantic similarity analysis using the CLIP model~\cite{icml/RadfordKHRGASAM21}. As shown in Figs.~\ref{fig:semantic_sim_c} and \ref{fig:semantic_sim_d}, the semantic similarity between the images and the corresponding category text (e.g., \textit{a photo of a \{dog\}}) is significantly improved when using our method. This suggests that our model can effectively disentangle and represent the semantic information within the latent space, leading to more semantically meaningful and interpretable features. The results of this analysis are consistent with the quantitative evaluation, further validating the effectiveness of our method.

\subsection{Ablation study}
In this section, we begin with ablation studies to examine the effect of code dimension and the role of uniformity regularization within SQ-GAN. We then delve into additional factors that could influence overall performance. Specifically, we document the usage of the codebook during the inference phase (i.e., the proportion of codebook entries actually utilized by the quantized vectors) to indicate how efficiently the codebook is being utilized and the extent of effective feature compression. Unless otherwise specified, all the ablation experiments are conducted on the Oxford-Dog dataset ~\cite{cvpr/ParkhiVZJ12}.

\begin{figure}[tbp]
\centering
\begin{subfigure}{0.48\linewidth}
\includegraphics[width=\linewidth]{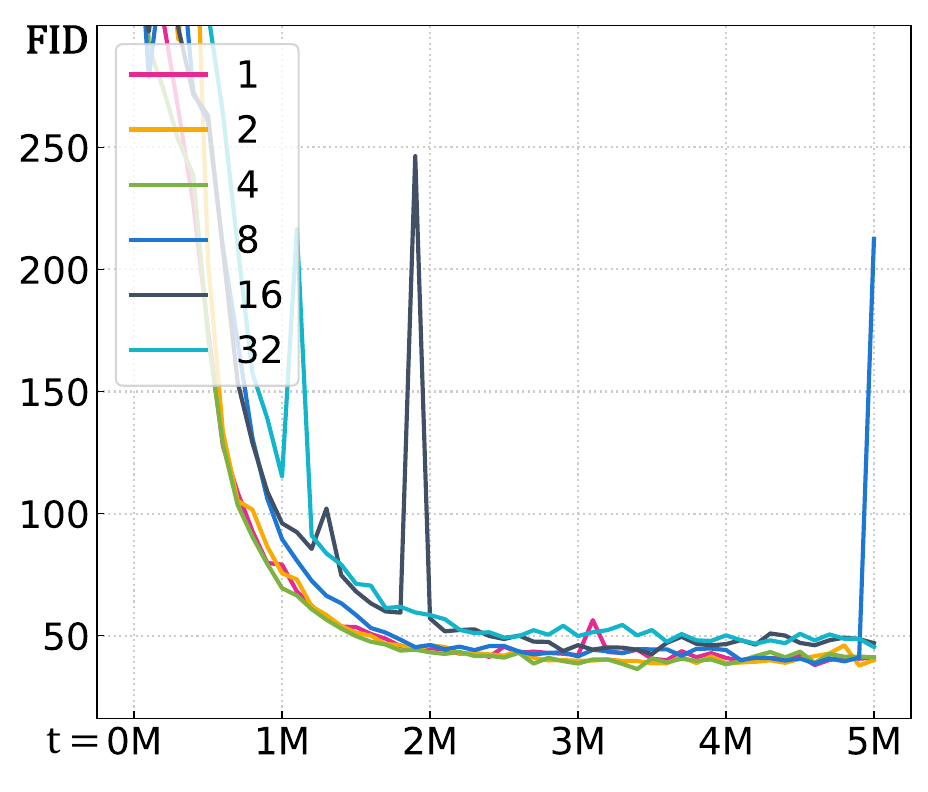}
\caption{Oxford-Dog}
\label{fig:fid_distribution_dog_dim}
\end{subfigure}
\begin{subfigure}{0.48\linewidth}
\includegraphics[width=\linewidth]{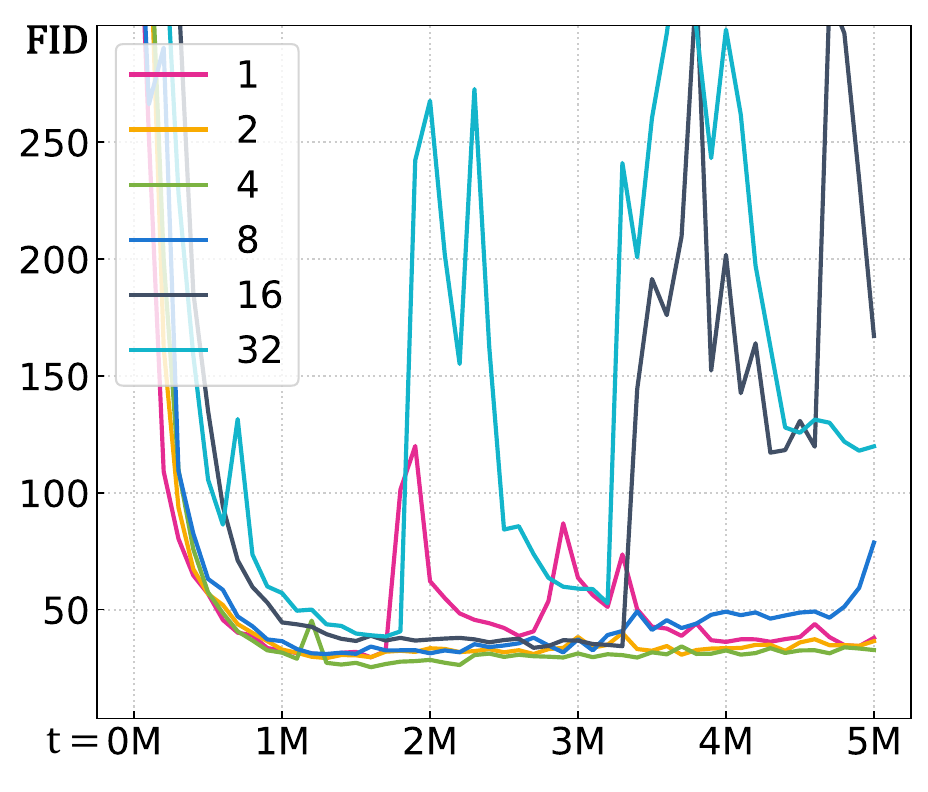}
\caption{FFHQ-2.5K}
\label{fig:fid_distribution_ffhq_dim}
\end{subfigure}
\vspace{-0.1in}
\caption{Evolutions of FID ($\Downarrow$) scores during training on the Oxford-Dog and FFHQ-2.5K datasets with different code dimensions.}
\label{fig:fid_distribution_dim}
\vspace{-0.1in}
\end{figure}

\begin{table}[tbp]
\centering
\resizebox*{\linewidth}{!}{
\begin{tabular}{|l|cccccc|c|c|r|rr|}
\hline
\multicolumn{1}{|c|}{\multirow{2}{*}{\textbf{Config.}}} & \multicolumn{6}{c|}{\textbf{Code dimension}} & \multirow{2}{*}{\textbf{Uniformity}} & \multirow{2}{*}{\textbf{Initialization}} & \multicolumn{1}{c|}{\multirow{2}{*}{\textbf{Usage(\%)}}} & \multicolumn{2}{c|}{\textbf{Oxford-Dog}}           \\
\cline{11-12} 
\multicolumn{1}{|c|}{} & 1 & 2 & 4 & 8 & 16 & 32 & & & \multicolumn{1}{c|}{} & \multicolumn{1}{c|}{FID $\Downarrow$} & \multicolumn{1}{c|}{IS $\Uparrow$} \\ 
\hline
\multirow{6}{*}{} & \Checkmark &    &    &    &     & & \Checkmark  & \XSolidBrush & 58.98 & 37.93 & 11.23 \\
&   & \Checkmark  &    &    &     & & \Checkmark  & \XSolidBrush & 75.50 & 37.71 & 11.06 \\
&   &    & \Checkmark  &    &     & & \Checkmark  & \XSolidBrush & 76.63 & 36.30 & 11.02 \\
&   &    &    & \Checkmark  &     & & \Checkmark  & \XSolidBrush & 69.18 & 38.60 & 11.02 \\
&   &    &    &    & \Checkmark   & & \Checkmark  & \XSolidBrush & 72.11 & 42.45 & 11.26 \\
&   &    &    &    &    & \Checkmark & \Checkmark  & \XSolidBrush & 71.01 & 45.33 & 11.01 \\
\cline{2-12} 
&   &    & \Checkmark  &    &     & & \XSolidBrush  & \XSolidBrush & 69.90 & 40.17 & 10.98 \\
\cline{2-11} 
\rowcolor{mygray} SQ-GAN &   &    & \Checkmark  &    &     & & \Checkmark  & \XSolidBrush & 76.63 & 36.30 & 11.52 \\
\hline
\rowcolor{mygray} SQ-GAN + CBI &   &    & \Checkmark  & & & & \Checkmark & \Checkmark & 78.55 & \textbf{35.01} & \textbf{12.44} \\
\hline
\end{tabular}
}
\vspace{-0.1in}
\caption{Ablation Studies of variant code dimension, uniformity regularization and codebook initialization on the Oxford-Dog dataset.}
\label{tab:ablation_study}
\vspace{-0.2in}
\end{table}

\paragraph{Effects of Code Dimension}
In Fig.~\ref{fig:fid_distribution_dim} and Table~\ref{tab:ablation_study}, we incrementally increased the code dimension of the codebook ($d_{\bm{w}}/s$) in our SQ-GAN from 1 to 32. The results show that when the code dimension is set to 4, the model's performance significantly improves. However, further increasing the code dimension does not yield additional improvement. This suggests that a code dimension of 4 is sufficient to capture the essential features of the data distribution, while higher dimensions may introduce unnecessary complexity and stability issues. This finding is consistent with the results of our comprehensive comparison.

\paragraph{Effects of codebook uniformity}
As demonstrated in Table~\ref{tab:ablation_study}, the improvement in the usage of codebook entries is markedly greater compared to settings without uniformity regularization, indicating that the discrete features learned in the codebook are more compact and representative. Additionally, we observe enhanced performance metrics when uniformity regularization is employed, highlighting how uniformity contributes to improving the representational quality of the codebook.

\paragraph{Effects of codebook initialization}
In Table \ref{tab:ablation_study}, we conducted an ablation study by replacing this knowledge-based initialization (i.e., ``+ CBI'') with random initialization. The results show a significant reduction in codebook usage (compared to the "+ CBI" setup with a code dimension of 4), along with a decreased performance on FID. However, even with random initialization, the performance of the codebook remains competitive with other methods (as shown in Table \ref{tab:comparison_comprehensive}). This indicates that our approach generates a more robust and expressive codebook when the knowledge-enhanced initialization method is employed. A more detailed discussion and theoretical analysis are provided in Appendix \myrc{A}.

\section{Conclusion}

In this study, we introduced a style space quantization method, leveraging codebook learning to enhance consistency regularization within the discriminator, particularly in limited-data scenarios. Furthermore, we advocate a pioneering approach to codebook initialization that incorporates external semantic knowledge from foundation models, thereby crafting a vocabulary-rich and perceptually diverse codebook. Extensive experiments across various tasks and datasets demonstrate that our approach results in high-quality image generation under limited data conditions. The discrete proxy space $\mathcal{W}^q$ we constructed serves as an advanced abstraction of the original $\mathcal{W}$ space, offering a more compact representation for image generation. Consequently, this paves the way for future research on stylized image generation, editing, and studies on attribute disentanglement and interpretability within the $\mathcal{W}$ space.

\section*{Acknowledgements}

This work is supported by the Fundamental Research Funds for the Central Universities under Grant 1082204112364, the National Major Scientific Instruments and Equipments Development Project of National Natural Science Foundation of China under Grant 62427820 and the Science Fund for Creative Research Groups of Sichuan Province Natural Science Foundation (No. 2024NSFTD0035).
{
    \small
    \bibliographystyle{ieeenat_fullname}
    \bibliography{main}
}

\appendix

\section{Analysis and discussion}

\subsection{Sinkhorn divergence algorithm}
The optimal transport (OT) problem is formally defined as follows:
\begin{equation}
\label{eq:naive_ot}
\begin{aligned}
D(\bm{p}, \bm{q} | \textbf{C}) &= \min_{\gamma \in \mathbb{R}^{n \times m}} \left<\gamma, \textbf{C}\right> \\
&= \min_{\gamma \in \mathbb{R}^{n \times m}} \sum_{i=1}^n \sum_{j=1}^m \gamma_{ij} c(\textbf{f}^x_i, \textbf{f}^y_j), \\
\text{subject to} \quad &\sum_{j=1}^m \gamma_{ij} = p_i \quad \forall i \in \{1, \ldots, n\}, \\
&\sum_{i=1}^n \gamma_{ij} = q_j \quad \forall j \in \{1, \ldots, m\},
\end{aligned}
\end{equation}
where $c(\textbf{f}^x_i, \textbf{f}^y_j)$ represents the cost of transporting a unit of mass from feature $\textbf{f}^x_i$ to $\textbf{f}^y_j$, and $\left<\gamma, \textbf{C}\right>$ denotes the Frobenius inner product of $\gamma$ and the cost matrix $\textbf{C}$.

Solving this optimization problem directly can be computationally prohibitive, especially in high-dimensional feature spaces. To mitigate this issue, we incorporate entropic regularization, which leads to the Sinkhorn distance and smooths the optimization landscape. The regularized OT problem is formulated as:
\begin{equation}
\begin{aligned}
D_{\epsilon}(\textbf{p}, \textbf{q} | \textbf{C}) &= \min_{\gamma \in \mathbb{R}^{n \times m}} \left<\gamma, \textbf{C}\right> - \epsilon h(\gamma), \\
\text{subject to} \quad &\gamma \mathbf{1}_m = \textbf{p}, \quad \gamma^\top \mathbf{1}_n = \textbf{q}, \\
&\gamma \in \mathbb{R}^{n \times m}_+,
\end{aligned}
\label{eq:reg_ot}
\end{equation}
where $h(\gamma) = - \sum_{i,j} \gamma_{ij} \log \gamma_{ij}$ is the entropy of $\gamma$, and $\mathbf{1_m}$, $\mathbf{1_n}$ are all-ones vectors.
The Sinkhorn distance allows for more efficient optimization through the iterative Sinkhorn-Knopp algorithm, where the optimal transport plan $\gamma$ is updated iteratively as:
\begin{equation}
\gamma^{(t)} = \text{diag}(\textbf{u}^{(t)}) \textbf{K} \text{diag}(\textbf{v}^{(t)}),
\end{equation}
where $\textbf{K} = \exp(-\textbf{C}/\epsilon)$ and $\textbf{u}$ and $\textbf{v}$ are updated as:

\begin{equation}
\begin{aligned}
\textbf{u}^{(t+1)} &= \textbf{p} \oslash (\textbf{K} \textbf{v}^{(t)}), \\
\textbf{v}^{(t+1)} &= \textbf{q} \oslash (\textbf{K}^\top \textbf{u}^{(t+1)}),
\end{aligned}
\end{equation}
with $\oslash$ denoting element-wise division and the initial condition $\textbf{v}^{(0)} = \mathbf{1}$.

\subsection{Detailed architecture design}

\begin{figure}[tbp]
\centering
\includegraphics[width=\linewidth]{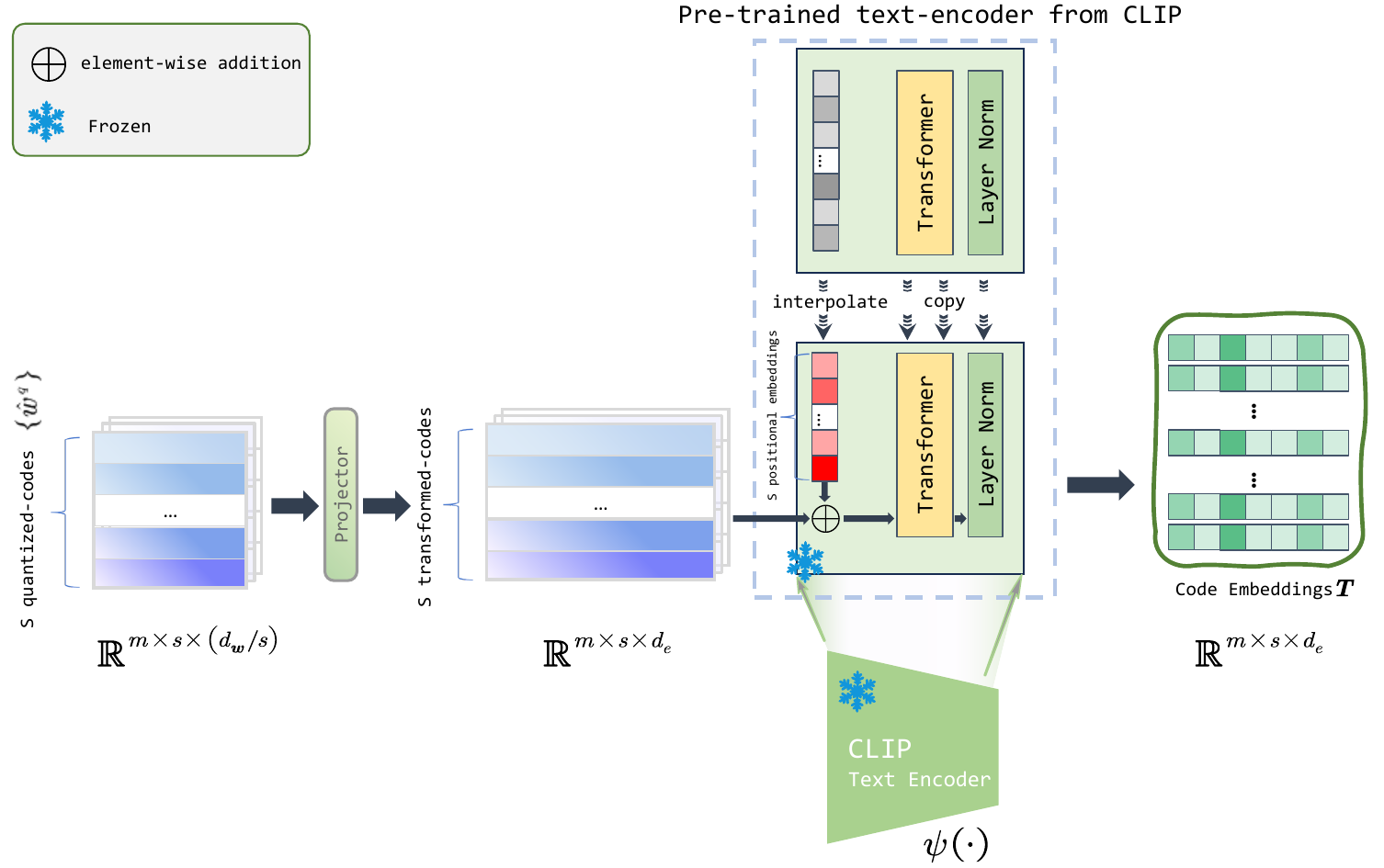}
\caption{Modifications to the positional embedding of CLIP's text encoder.}
\label{fig:modified_PE}
\vspace{-0.1in}
\end{figure}

To ensure compatibility with the Transformer architecture in our approach, the quantized sub-vectors $\{\hat{\bm{w}}^q_i\}_{i=1}^m \subset \mathbb{R}^{m \times s \times (d_{\bm{w}}/s)}$ are processed through a learnable MLP layer, which adjusts them to match the embedding dimension of the input tokens. These transformed variables are then input into a Transformer model, utilizing the CLIP~\cite{icml/RadfordKHRGASAM21} text encoder. In the original CLIP model, absolute positional embeddings are added to the token embeddings to encode positional information within the sequence. As depicted in Fig.~\ref{fig:modified_PE}, to align the pre-trained positional embeddings with the size of our transformed variables, We modify the context length in the CLIP model to correspond to the number of quantized sub-vectors, denoted by $s$, and interpolate the pre-trained positional embeddings to match this new length. This adjustment also necessitates corresponding changes in the attention mask within the Transformer, ensuring proper functionality.

\subsection{Usage of codebook}
Our primary objective in this work is to learn a compact and thoroughly explored latent space. To achieve this, we aim to maximize the activation of codes within the codebook. The utilization rate of the codebook is defined as the proportion of active codes. Our results (Table~\myrc{4} in primary text) show that smaller code dimensions are more effectively utilized, likely due to the simplicity and decoupling of the information they encode. Furthermore, our codebook initialization method significantly enhances the utilization rate, indicating that embedded prior knowledge facilitates better use of the codebook information.

\begin{figure}[!tbp]
\centering
\begin{subfigure}{0.473\linewidth}
\includegraphics[width=\linewidth]{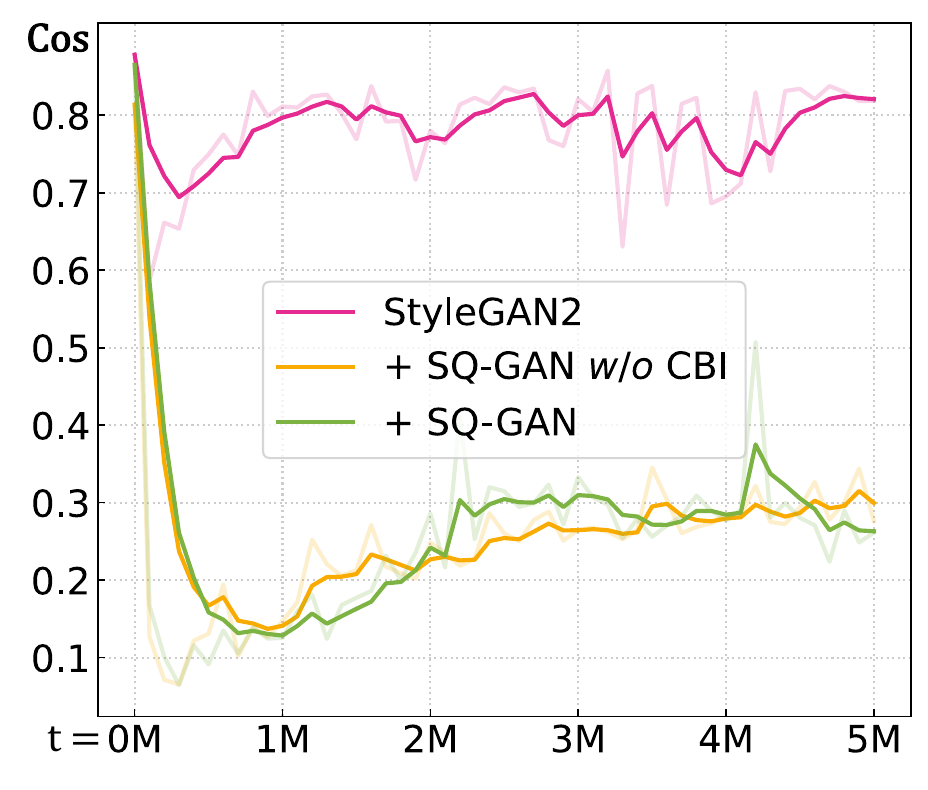}
\caption{Discriminator}
\label{fig:sim_a}
\end{subfigure}
\begin{subfigure}{0.495\linewidth}
\includegraphics[width=\linewidth]{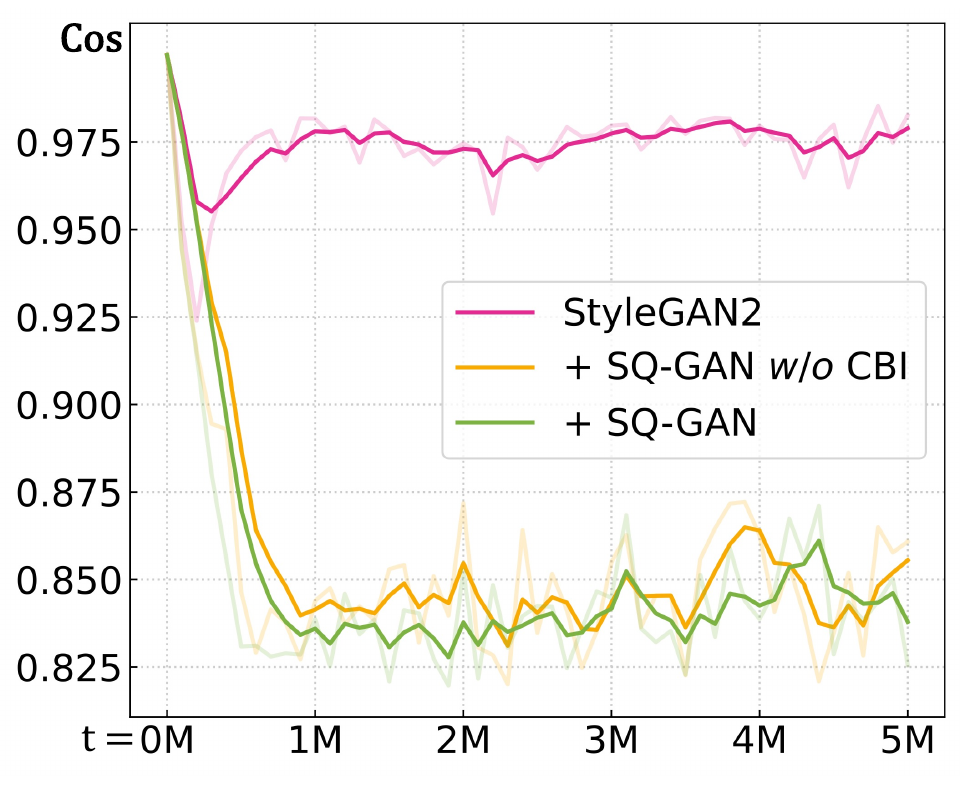}
\caption{CLIP}
\label{fig:sim_b}
\end{subfigure}
\caption{Evolutions of the cosine similarity for the discriminator's embedding space and CLIP's embedding space.}
\label{fig:sim}
\vspace{-0.1in}
\end{figure}

\subsection{Embedding space}
We evaluate the diversity within the discriminator's embedding space by analyzing the similarity between extracted features. Feature vectors are drawn from the discriminator's penultimate layer, and cosine similarity is computed across all pairs of feature vectors in the dataset. By tracking the average cosine similarity throughout training, we gain insights into the evolution of the embedding space. Fig.~\ref{fig:sim_a} illustrates the cosine similarity's progression in the discriminator's embedding space. Compared to the baseline method, our approach demonstrates a lower average cosine similarity, indicating a more diverse and discriminative embedding space.

Additionally, we evaluate the similarity between the CLIP model's features on the generated images, as shown in Fig.~\ref{fig:sim_b}. The results reveal that our method achieves a lower average cosine similarity, indicating greater diversity in the generated images, aligning with our approach's objectives.

\section{Experimental settings}

\subsection{Datasets}
Our experiments are conducted using four distinct datasets: Oxford-Dog (derived from the Oxford-IIIT Pet Dataset~\cite{cvpr/ParkhiVZJ12}), Flickr-Faces-HQ (FFHQ)~\cite{cvpr/KarrasLA19}, MetFaces~\cite{nips/KarrasAHLLA20}, and BreCaHAD~\cite{brn/Aksac2019}. Detailed descriptions of each dataset are provided below:

\subsubsection{OxfordDog}
We utilize dog images from the Oxford-IIIT Pet Dataset~\cite{cvpr/ParkhiVZJ12}. A dog face detection model is employed to crop and standardize these images to a uniform resolution of $256 \times 256$ pixels, ensuring the dog's face is centered as accurately as possible. A total of 4,492 images are randomly selected for training, while the remaining 498 images constitute the test set. This dataset includes approximately 25 dog breeds, presenting a challenging variety of poses and backgrounds, thereby offering a diverse set of conditions for our experiments.

\subsubsection{FFHQ}
The Flickr-Faces-HQ (FFHQ) dataset~\cite{cvpr/KarrasLA19} comprises 70,000 high-resolution images of human faces, representing a broad spectrum of ages, ethnicities, and backgrounds. These images, sourced from Flickr, have been meticulously aligned and cropped to ensure high consistency and quality across the dataset. The dataset also includes various accessories such as eyeglasses and hats, further enhancing its diversity.

\subsubsection{MetFaces}
The MetFaces dataset~\cite{nips/KarrasAHLLA20} contains 1,336 high-resolution images of faces from the Metropolitan Museum of Art's collection\footnote{https://metmuseum.github.io/}. These images are used primarily for research and analysis in facial representations across different artistic styles, making this dataset a unique resource for evaluating generative models.

\subsubsection{BreCaHAD}
The BreCaHAD dataset~\cite{brn/Aksac2019} is specifically designed for breast cancer histopathology research. It contains 162 high-resolution images ($1360 \times 1024$ pixels) of histopathology slides. For our experiments, these images are restructured into 1,944 partially overlapping crops, each with a resolution of $512 \times 512$ pixels.

\subsubsection{FFHQ-2.5k}
\begin{figure}[!htbp]
\centering
\includegraphics[width=0.9\linewidth]{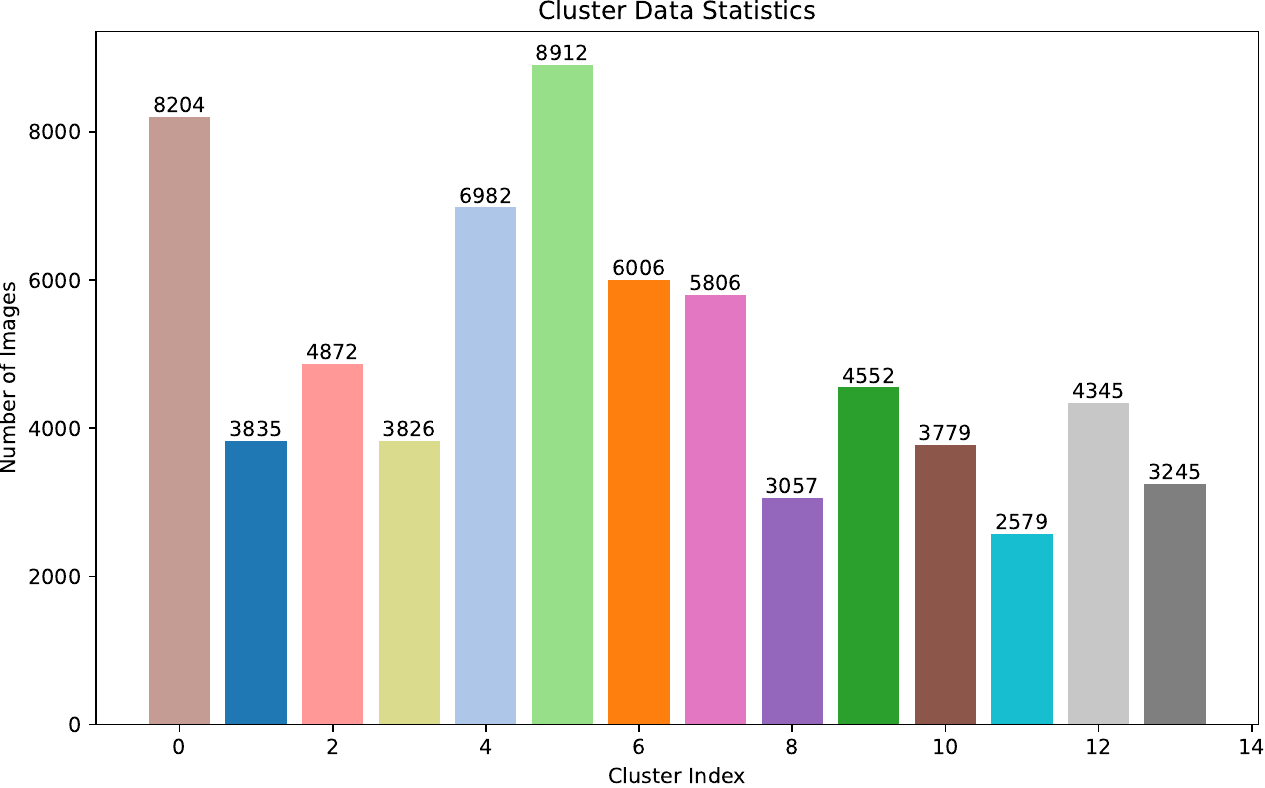}
\caption{Clustering results for the FFHQ dataset.}
\label{fig:ffhq-all_polar}
\vspace{-0.1in}
\end{figure}

\begin{figure}[!htbp]
\centering
\includegraphics[width=0.9\linewidth]{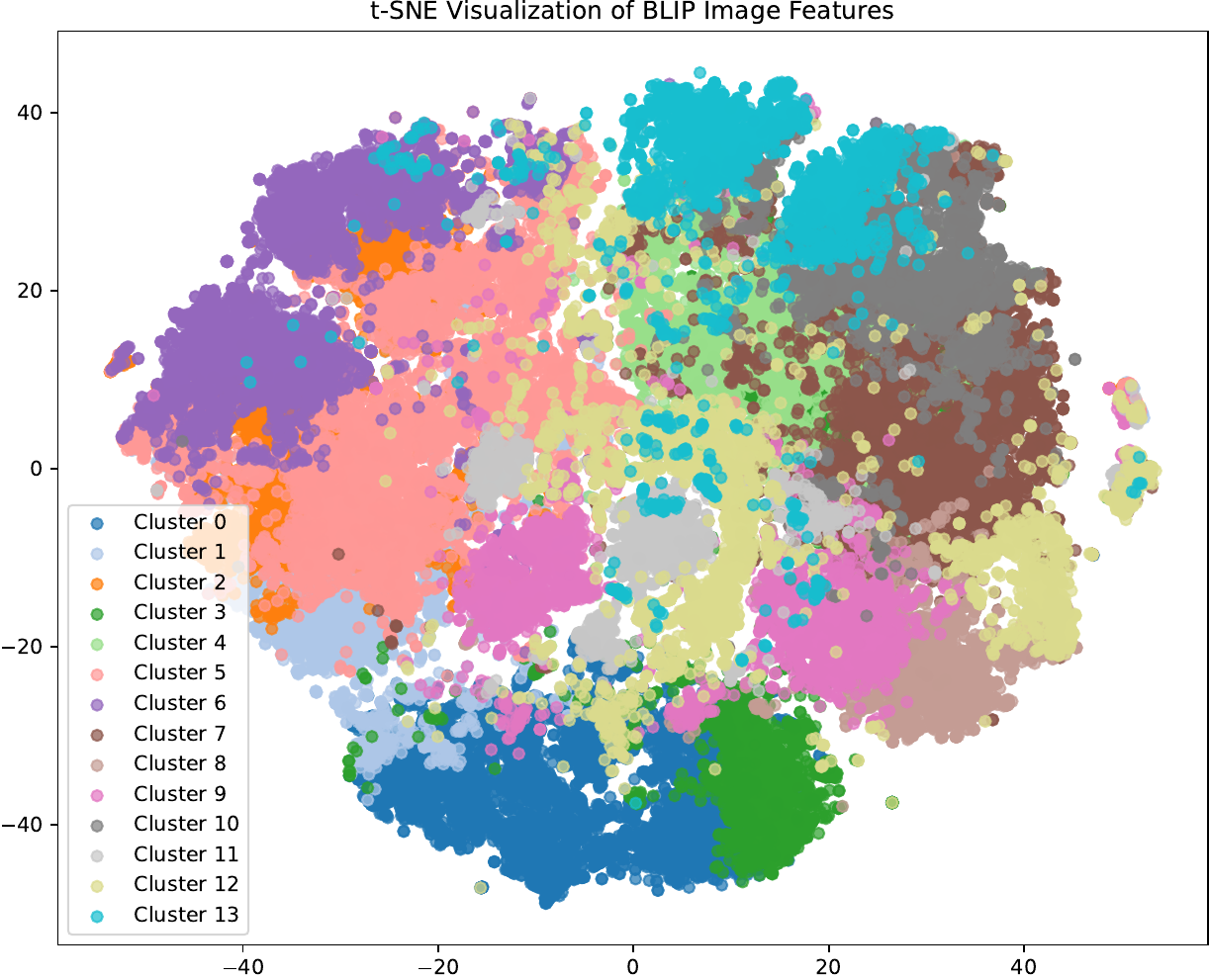}
\caption{t-SNE visualization of features from the FFHQ dataset.}
\label{fig:ffhq-all_t-SNE}
\vspace{-0.1in}
\end{figure}

We apply a pre-trained BLIP-base model~\cite{icml/0008LSH23} to the original 70,000 images from the FFHQ dataset to extract features. These features are then aggregated to facilitate the application of the K-means clustering algorithm. To simulate a low-data scenario, we set the number of cluster centers to K=14, resulting in an average of 5,000 images per cluster. We present the distribution of these clusters (Fig.~\ref{fig:ffhq-all_polar}) and visualize the features using t-SNE (Fig.~\ref{fig:ffhq-all_t-SNE}). From these clusters, we select the smallest, comprising 2,500 images (referred to as FFHQ-2.5K), for further experimentation.

\subsection{Evaluation metrics} 

To evaluate the efficacy of our proposed method and benchmark it against existing baselines, we employ three widely recognized metrics: Inception Score (IS)~\cite{nips/SalimansGZCRCC16}, Fréchet Inception Distance (FID)~\cite{nips/HeuselRUNH17}, and Kernel Inception Distance (KID)~\cite{iclr/BinkowskiSAG18}. These metrics provide a comprehensive assessment of the quality, diversity, and distribution alignment of the generated images, enabling a thorough comparison of the models' performance. The following sections provide a detailed overview of each metric:

\subsubsection{Inception Score (IS)}
The IS metric~\cite{nips/SalimansGZCRCC16} evaluates both the quality and diversity of the generated images. It is a widely recognized measure in the early stages of GAN development, which evaluates the generated images by analyzing the conditional entropy of class labels predicted by an Inception network, with a higher score indicating better performance. 

\subsubsection{Fréchet Inception Distance (FID)}
The FID metric~\cite{nips/HeuselRUNH17} is extensively used to measure the similarity between the distributions of generated and real images. It computes the Fréchet distance—also known as the Wasserstein-2 distance, between Gaussian distributions fitted to the hidden activations of an Inception network for both the generated and ground-truth images. FID is sensitive to both the quality and diversity of the images, with a lower score reflecting superior performance. It is considered a more comprehensive and reliable indicator than IS, particularly in capturing discrepancies in higher-order statistics.

\subsubsection{Kernel Inception Distance (KID)}
The KID metric~\cite{iclr/BinkowskiSAG18} is similar to FID but offers distinct advantages. It measures the squared maximum mean discrepancy (MMD) between Inception representations of the generated and real images. Unlike FID, KID does not assume a parametric form for the activation distribution, and it provides a simple, unbiased estimator. This makes KID especially informative when the available ground-truth data is limited in scale. A lower KID score indicates better alignment between the generated and real data distributions, signaling higher image quality and consistency.

\subsection{Implementation details}

We utilized the implementations from~\cite{iclr/MiyatoKKY18} and ~\cite{cvpr/KarrasLAHLA20} to train SNGAN and StyleGAN2, respectively. Additionally, we implemented a $64 \times 64$ version of DCGAN based on the approach outlined in~\cite{corr/RadfordMC15}. All experiments were conducted using consistent hyperparameter settings across models, and performance was evaluated using the evaluation framework provided by~\cite{cvpr/KarrasLAHLA20}.
During training, we adjusted the settings for $\lambda_{\text{sq}}$ and $\lambda_{\text{qcr}}$ to 0.01 each, reflecting the additional constraint terms introduced by StyleGAN2. It is worth noting that our results exhibited some discrepancies when compared to the scores reported in the literature, which may be attributed to variations in hardware or differences across experimental runs.

\section{Additional Results}
\begin{figure*}
\setlength{\tabcolsep}{1pt}
\centering
\includegraphics[width=\linewidth]{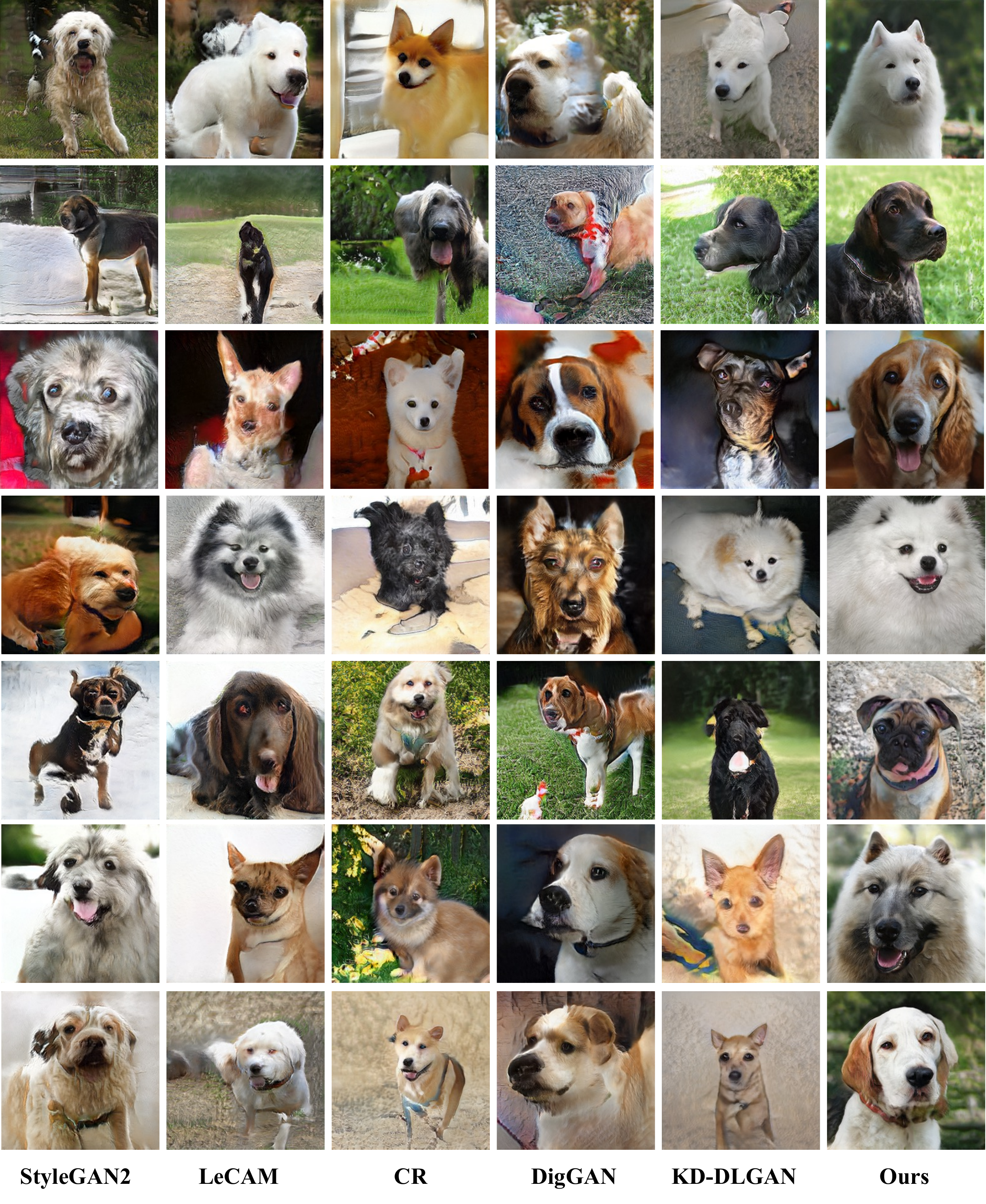
}
\caption{Quantitative results on the OxfordDog dataset (best FID).}
\vspace{-0.25cm}
\label{fig:OxfordDog_quantitative}
\end{figure*}

\begin{figure*}
\setlength{\tabcolsep}{1pt}
\centering
\includegraphics[width=\linewidth]{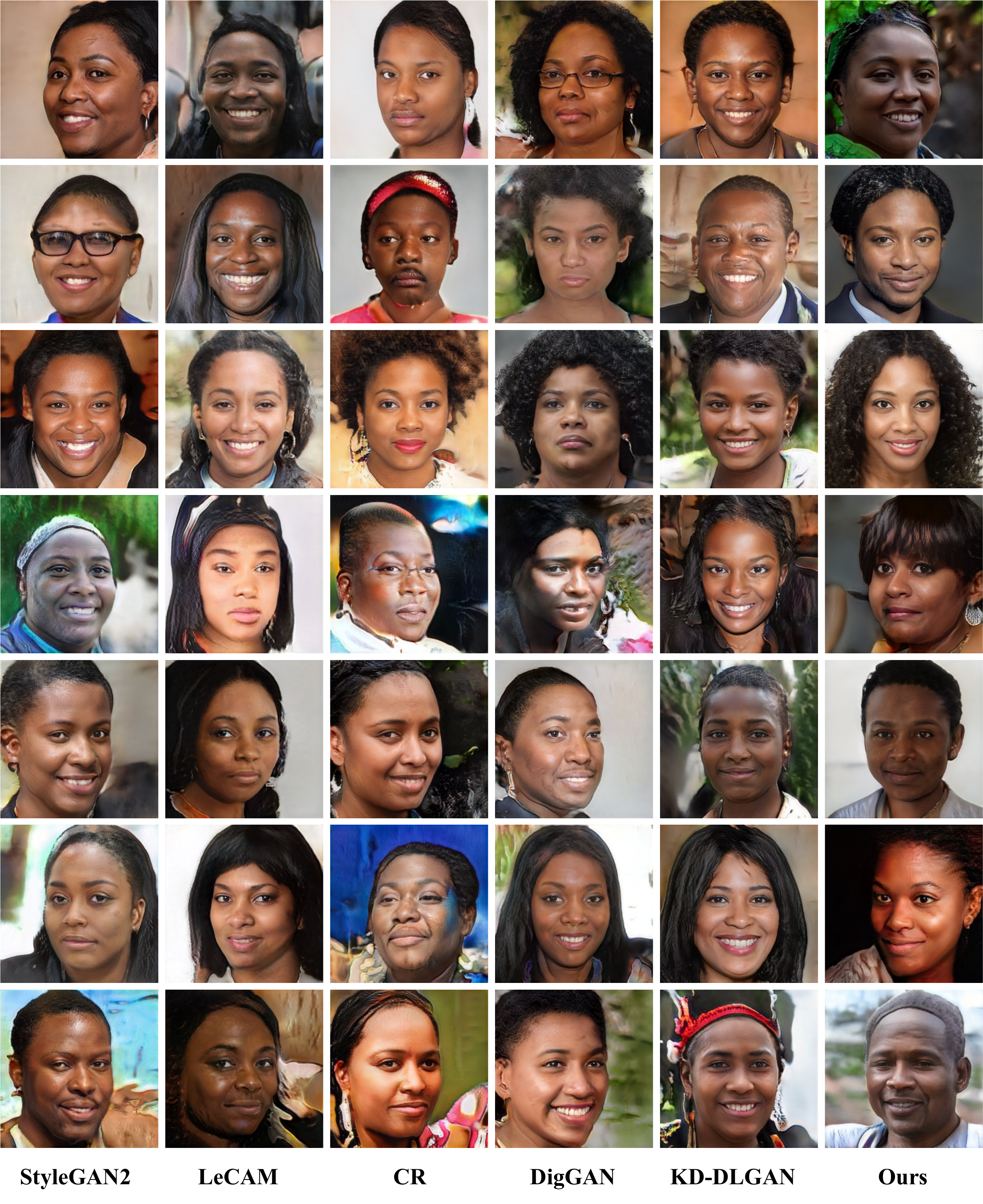
}
\caption{Quantitative results on the FFHQ-2.5k dataset (best FID).}
\vspace{-0.25cm}
\label{fig:FFHQ_quantitative}
\end{figure*}

\begin{figure*}
\setlength{\tabcolsep}{1pt}
\centering
\includegraphics[width=\linewidth]{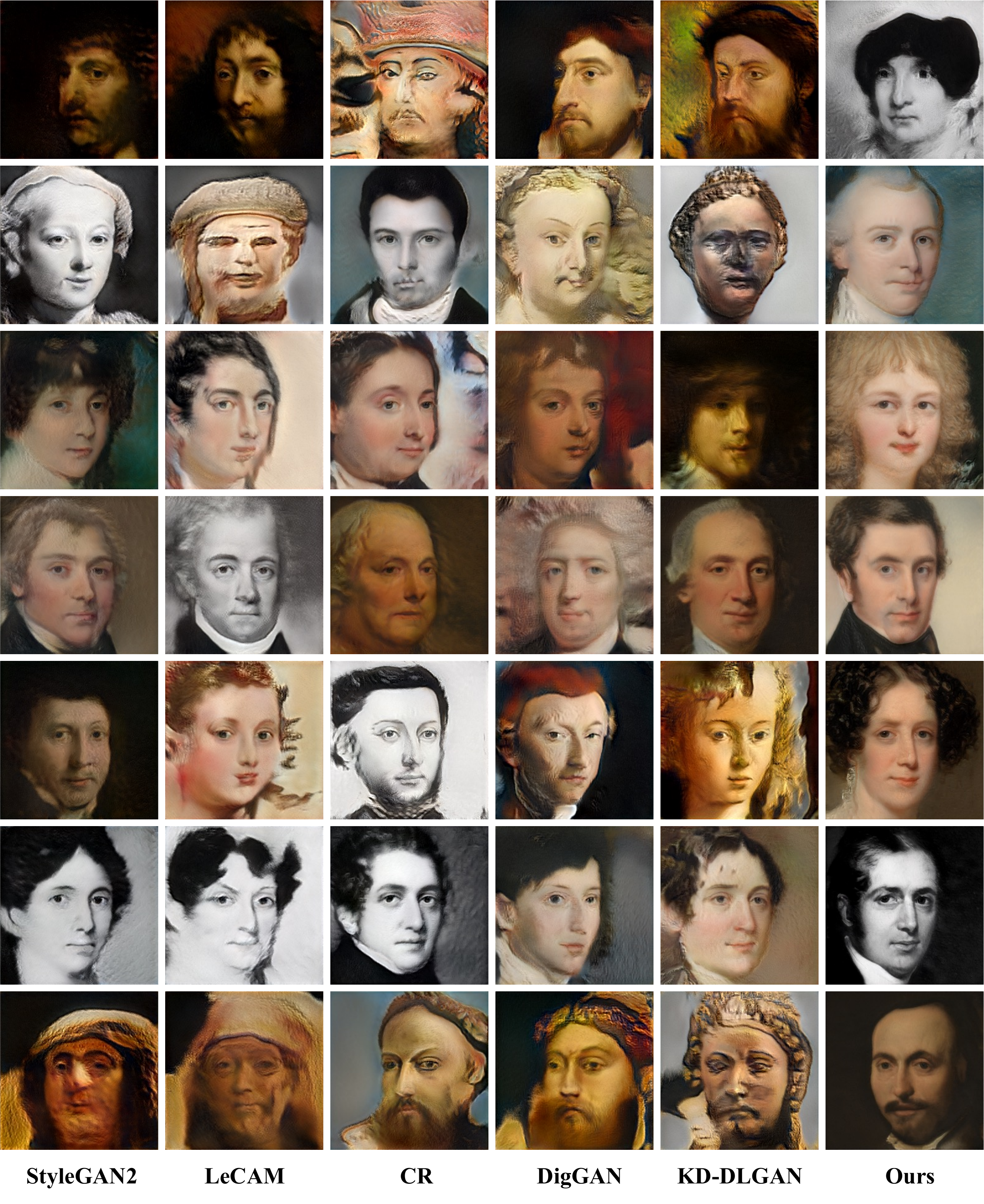
}
\caption{Quantitative results on the MetFaces dataset (best FID).}
\vspace{-0.25cm}
\label{fig:metfaces_quantitative}
\end{figure*}

\begin{figure*}
\setlength{\tabcolsep}{1pt}
\centering
\includegraphics[width=\linewidth]{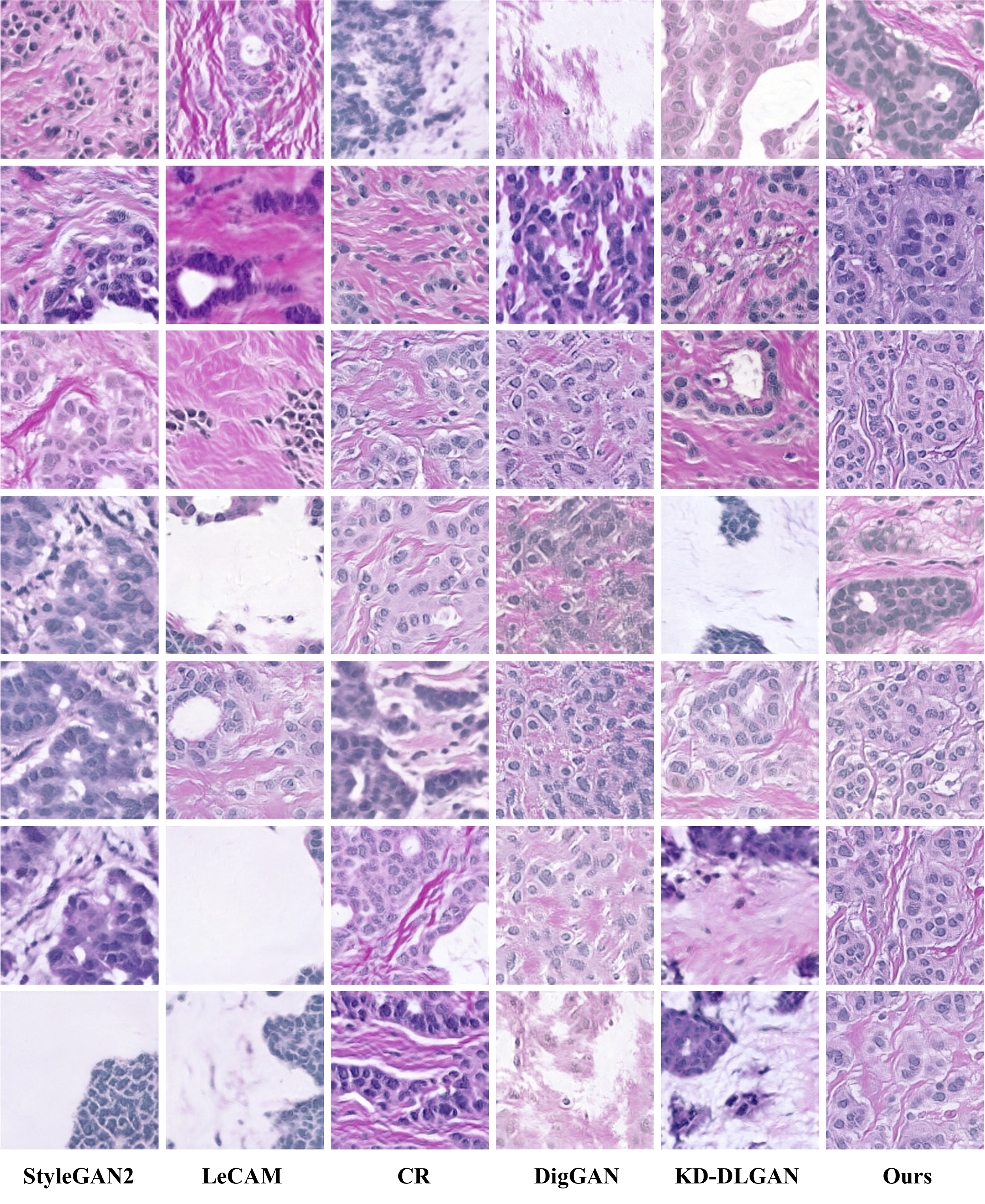
}
\caption{Quantitative results on the BreCaHAD dataset (best FID).}
\vspace{-0.25cm}
\label{fig:brecahad_quantitative}
\end{figure*}

The superior performance of our method is further validated by the qualitative results presented in Figs. \ref{fig:OxfordDog_quantitative}, \ref{fig:FFHQ_quantitative}, \ref{fig:metfaces_quantitative} and \ref{fig:brecahad_quantitative}, which showcase the high-quality images generated by our approach. For consistency, identical hyperparameters and the same random seed were maintained across all experiments. The images presented were randomly selected from the generated outputs, with no specific selection criteria other than a global random seed. ``Best FID'' refers to images generated at the step with the best FID score. Our results indicate that our method produces more realistic images compared to baseline models.

\end{document}